\documentclass[a4paper]{article}

\usepackage{iclr2016_conference,times}
\usepackage{hyperref}
\usepackage{url}
\usepackage[english]{babel}
\usepackage[utf8]{inputenc}
\usepackage{algorithm}
\usepackage{algorithmic}
\usepackage{amsfonts}
\usepackage{amsmath}
\usepackage{amssymb}
\usepackage{bm}
\usepackage{multirow}
\usepackage[super]{nth}
\usepackage{graphicx}
\usepackage[colorinlistoftodos]{todonotes}

\iclrfinalcopy
\newcommand{\atariscore}[1]{\text{score}_\text{#1}}

\title{Prioritized Experience Replay}

\author{Tom Schaul, John Quan, Ioannis Antonoglou and David Silver\\
Google DeepMind\\
\texttt{\{schaul,johnquan,ioannisa,davidsilver\}@google.com} \\
}

\begin{document}
\maketitle

\begin{abstract}
Experience replay lets online reinforcement learning agents remember and reuse experiences from the past. In prior work, experience transitions were uniformly sampled from a replay memory. However, this approach simply replays transitions at the same frequency that they were originally experienced, regardless of their significance. In this paper we develop a framework for prioritizing experience, so as to replay important transitions more frequently, and therefore learn more efficiently. We use prioritized experience replay in Deep Q-Networks (DQN), a reinforcement learning algorithm that achieved human-level performance across many Atari games. DQN with prioritized experience replay achieves a new state-of-the-art, outperforming DQN with uniform replay on 41 out of 49 games.
\end{abstract}

\section{Introduction}
Online reinforcement learning (RL) agents incrementally update their parameters
(of the policy, value function or model) while they observe a stream of
experience. In their simplest form, they discard incoming data immediately, after a single update.
Two issues with this are (a) strongly correlated updates that break the i.i.d. assumption 
of many popular stochastic gradient-based algorithms, and (b) the rapid forgetting of possibly
rare experiences that would be useful later on.

\emph{Experience replay}~\citep{lin-er} addresses both of these issues: with experience stored in a replay memory,
it becomes possible to break the temporal correlations by mixing more and less recent experience for the updates,
and rare experience will be used for more than just a single update.
This was demonstrated in the Deep Q-Network (DQN) algorithm~\citep{dqn-workshop,dqn-nature}, which stabilized the training of
a value function, represented by a deep neural network, by using experience replay. Specifically, DQN used a large sliding window replay memory, sampled from it uniformly at random, and revisited each transition%
\footnote{
A transition is the atomic unit of interaction in RL, in our case a tuple of 
(state $S_{t-1}$, action $A_{t-1}$, reward $R_{t}$, discount $\gamma_{t}$, next state $S_{t}$). 
We choose this for simplicity, but most of the arguments in this paper also hold for a coarser ways of chunking experience, e.g.\ into sequences or episodes.
}
eight times on average.
In general, experience replay can reduce the amount of experience required to learn,
and replace it with more computation and more memory -- which are often cheaper resources than the RL agent's interactions with its environment.

In this paper, we investigate how \emph{prioritizing} which transitions are replayed 
can make experience replay more efficient and effective than if all transitions are replayed uniformly.
The key idea is that an RL agent can learn more effectively from some transitions than from others.
Transitions may be more or less surprising, redundant, or task-relevant. 
Some transitions may not be immediately useful to the agent, but might become so when the agent competence increases~\citep{curiosity}.
Experience replay liberates online learning agents from processing transitions 
in the exact order they are experienced. 
Prioritized replay further liberates agents from considering transitions with the same frequency that they are experienced.

In particular, we propose to more frequently replay transitions with high expected learning progress, 
as measured by the magnitude of their temporal-difference (TD) error. 
This prioritization can lead to a loss of diversity, which we alleviate with stochastic prioritization,
and introduce bias, which we correct with importance sampling.
Our resulting algorithms are robust and scalable, which we demonstrate
on the Atari 2600 benchmark suite, where we obtain faster learning and state-of-the-art performance.

\section{Background}

Numerous neuroscience studies have identified evidence of experience replay in the hippocampus of rodents, suggesting that sequences of prior experience are replayed, either during awake resting or sleep. Sequences associated with rewards appear to be replayed more frequently~\citep{dharsh1,dharsh2,dharsh3}. Experiences with high magnitude TD error also appear to be replayed more often~\citep{dharsh4,dharsh5}. 

It is well-known that planning algorithms such as value iteration can be made more efficient by prioritizing updates in an appropriate order. \emph{Prioritized sweeping} ~\citep{prioritized-sweeping,gen-sweeping} selects which state to update next, prioritized according to the change in value, if that update was executed. The TD error provides one way to measure these priorities~\citep{harm}. Our approach uses a similar prioritization method, but for model-free RL rather than model-based planning. Furthermore, we use a stochastic prioritization that is more robust when learning a function approximator from samples.

TD-errors have also been used as a prioritization mechanism for determining where to focus resources, for example when choosing where to explore~\citep{white2014surprise} or which features to select~\citep{geramifard2011online,sun2011incremental}.

In supervised learning, there are numerous techniques to deal with imbalanced datasets
when class identities are known, including re-sampling, under-sampling and over-sampling techniques,
possibly combined with ensemble methods~\citep[for a review, see][]{imbalance-ensemble}.
A recent paper introduced a form of re-sampling in the context of deep RL with experience replay~\citep{narasimhan2015language}; the method separates experience into two buckets, one for positive and one for negative rewards, and then picks a fixed fraction from each to replay. This is only applicable in domains that (unlike ours) have a natural notion of `positive/negative' experience.
Furthermore, \citet{hinton2007} introduced a form of non-uniform sampling based on error, with an importance sampling correction,
which led to a 3x speed-up on MNIST digit classification.

There have been several proposed methods for playing Atari with deep reinforcement learning, including deep Q-networks (DQN)~\citep{dqn-workshop,dqn-nature,dagger,Stadie:2015,gorila,increasing_action_gap}, and the \emph{Double DQN} algorithm \citep{double-dqn}, which is the current published state-of-the-art.
Simultaneously with our work, an architectural innovation that separates advantages from the value function (see the co-submission by~\citealp{dueling}) 
has also led to substantial improvements on the Atari benchmark.

\section{Prioritized Replay}
\label{sec-preplay}

Using a replay memory leads to design choices at two levels: which experiences to store,
and which experiences to replay (and how to do so). This paper addresses only the latter: making the
most effective use of the replay memory for learning, assuming that its contents are outside of our control 
(but see also Section~\ref{sec-prio-memory}).

\subsection{A Motivating Example}
\label{sec:example}

\begin{figure}[tb]
\vspace{-1em}
\centerline{
\includegraphics[width=0.45\textwidth]{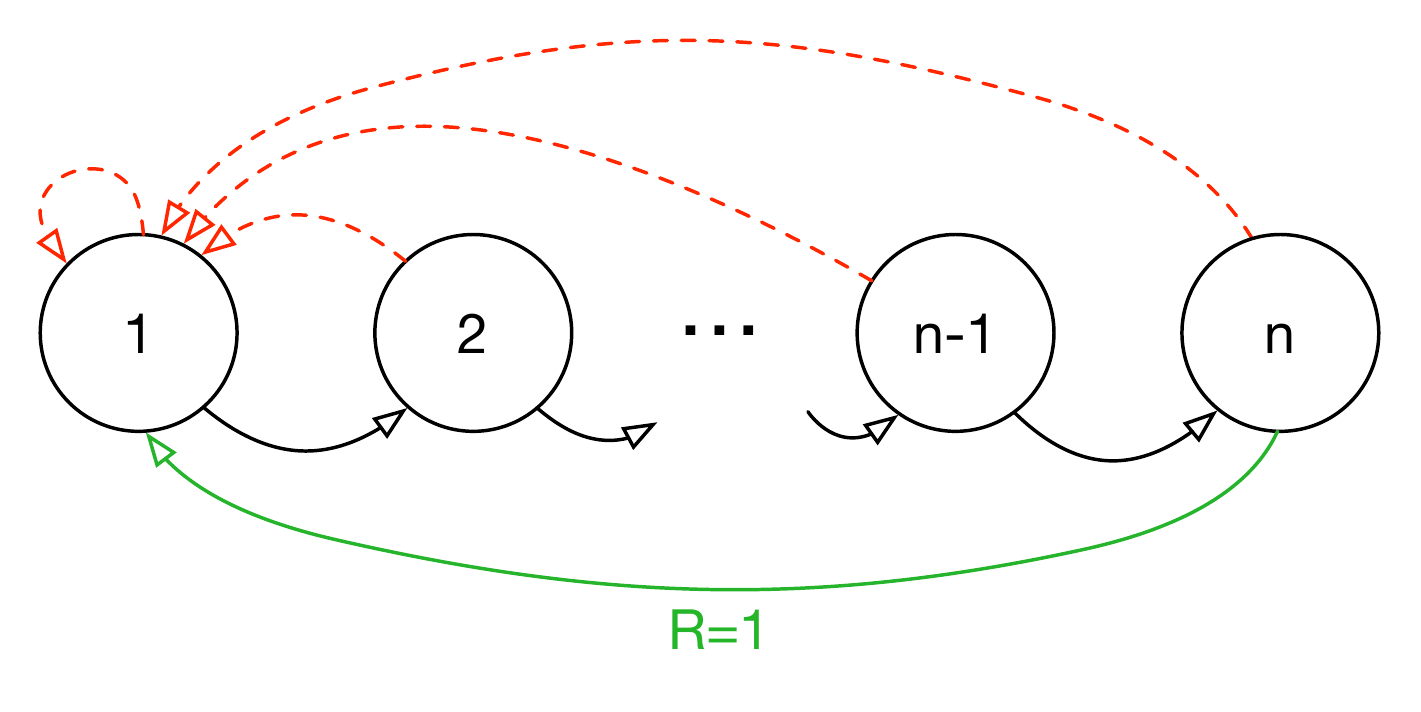}
\hspace{0.1\textwidth}
\includegraphics[width=0.4\textwidth]{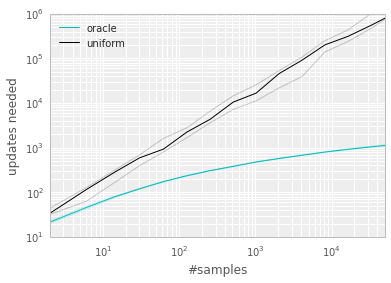}
}
\vspace{-0.5em}
\caption{
\label{fig-microzuma-chain}
{\bf Left}: Illustration of the `Blind Cliffwalk' example domain: there are two actions, a `right' and a `wrong' one, 
and the episode is terminated whenever the agent takes the `wrong' action (dashed red arrows). 
Taking the `right' action progresses through a sequence of $n$ states (black arrows), at the end of which lies
a final reward of $1$ (green arrow); reward is $0$ elsewhere. We chose a representation such that generalizing over what action is `right' is not possible.
{\bf Right}: Median number of learning steps required to learn the value function as a function of the size of the total number of transitions in the replay memory. 
Note the log-log scale, which highlights the exponential speed-up from replaying with an oracle (bright blue), compared to uniform replay (black); faint lines are min/max values from 10 independent runs.
\vspace{-1em}
}
\end{figure}

To understand the potential gains of prioritization, we introduce an artificial `Blind Cliffwalk' environment (described in Figure~\ref{fig-microzuma-chain}, left)
that exemplifies the challenge of exploration when rewards
are rare.
With only $n$ states, the environment requires an exponential number of random steps until the first non-zero reward; 
to be precise, the chance that a random sequence of actions will lead to the reward is $2^{-n}$.
Furthermore, the most relevant transitions (from rare successes) 
are hidden in a mass of highly redundant failure cases (similar to a bipedal robot falling over repeatedly, before it discovers how to walk).

We use this example to highlight the difference between the learning times of two agents. Both agents perform Q-learning updates on transitions drawn from the same replay memory. The first agent replays transitions uniformly at random, while the second agent invokes an oracle to prioritize transitions. This oracle greedily selects the transition that maximally reduces the global loss in its current state (in hindsight, after the parameter update).
For the details of the setup, see Appendix~\ref{app-microzuma}.
Figure~\ref{fig-microzuma-chain} (right) shows that picking the transitions in a good order
can lead to exponential speed-ups over uniform choice.
Such an oracle is of course not realistic, yet the large gap motivates our search 
for a practical approach that improves on uniform random replay.

\subsection{Prioritizing with TD-error}
The central component of prioritized replay is the criterion by which the importance of each transition is measured. 
One idealised criterion would be the amount the RL agent
can learn from a transition in its current state (expected learning progress). 
While this measure is not directly accessible, a reasonable proxy is the magnitude of 
a transition's TD error $\delta$, which indicates how `surprising' or unexpected the transition is: specifically, how far the value is from its next-step bootstrap estimate~\citep{gen-sweeping}.
This is particularly suitable for incremental, online RL algorithms, such as SARSA or Q-learning, that already compute the TD-error and update
the parameters in proportion to $\delta$.
The TD-error can be a poor estimate in some circumstances as well, e.g.~when rewards are noisy; 
see Appendix~\ref{sec-alternative-td} for a discussion of alternatives.

\begin{figure}[tb]
\vspace{-0.5em}
\centerline{
\includegraphics[width=0.475\textwidth]{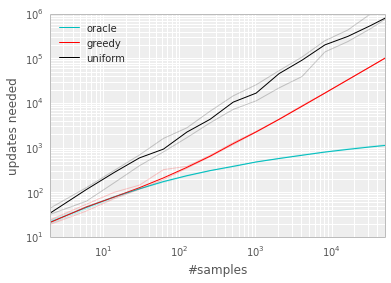}
\hspace{0.05\textwidth}
\includegraphics[width=0.475\textwidth]{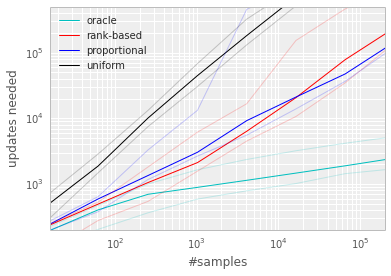}
}
\vspace{-0.5em}
\caption{
\label{fig-baseline-oracle}
Median number of updates required for Q-learning to learn the value function on the Blind Cliffwalk example, as a function of the total number of transitions (only a single one of which was successful and saw the non-zero reward). Faint lines are min/max values from 10 random initializations.
Black is uniform random replay, cyan uses the hindsight-oracle to select transitions, red and blue use prioritized replay (rank-based and proportional respectively).
The results differ by multiple orders of magnitude, thus the need for a log-log plot.
In both subplots it is evident that replaying experience in the right order makes an enormous difference to the number of updates required. See Appendix~\ref{app-microzuma} for details.
{\bf Left:} Tabular representation, greedy prioritization. {\bf Right}: Linear function approximation, both variants of stochastic prioritization.
\vspace{-1em}
}
\end{figure}

To demonstrate the potential effectiveness of prioritizing replay by TD error, 
we compare the uniform and oracle baselines in the Blind Cliffwalk to 
a `greedy TD-error prioritization' algorithm. This algorithm stores the last encountered TD error along with each transition in the replay memory. The transition with the largest absolute TD error is replayed from the memory. A Q-learning update is applied to this transition, which updates the weights in proportion to the TD error. 
New transitions arrive without a known TD-error, so we put them at maximal priority in order to guarantee that
all experience is seen at least once. 
Figure~\ref{fig-baseline-oracle} (left), shows that this algorithm results in a substantial reduction in the effort required to solve the Blind Cliffwalk task.%
\footnote{
Note that a random (or optimistic) initialization of the Q-values is necessary with greedy prioritization.
If initializing with zero instead, unrewarded transitions would appear to have zero error initially, be placed at the bottom of the queue, and not be revisited until the error on other transitions drops below numerical precision.
}

{\bf Implementation}: To scale to large memory sizes $N$,
we use a binary heap data structure for the priority queue, for which 
finding the maximum priority transition when sampling is $O(1)$
and updating priorities (with the new TD-error after a learning step) is $O(\log N)$.
See Appendix~\ref{app-atari-experiments-impl-details} for more details.

\subsection{Stochastic Prioritization}

However, greedy TD-error prioritization has several issues.
First, to avoid expensive sweeps over the entire replay memory, TD errors are only updated for the transitions that are replayed. One consequence is that transitions that have a low TD error on first visit may not be replayed for a long time 
(which means effectively never with a sliding window replay memory).
Further, it is sensitive to noise spikes (e.g.\ when rewards are stochastic), 
which can be exacerbated by bootstrapping, where approximation errors appear as another source of noise.
Finally, greedy prioritization focuses on a small subset of the experience: errors shrink slowly, especially when using function approximation, meaning that the initially high error transitions get replayed frequently. This lack of diversity that makes the system prone to over-fitting.

To overcome these issues, we introduce a stochastic sampling method that
interpolates between pure greedy prioritization and uniform random sampling. We ensure that the probability of being sampled is monotonic in a transition's priority, 
while guaranteeing a non-zero probability even for the lowest-priority transition.
Concretely, we define the probability of sampling transition $i$ as
\begin{equation}
\label{eq-distr}
P(i) = \frac{p_i^{\alpha}}{\sum_k p_k^{\alpha}}
\end{equation}
where $p_i > 0$ is the priority of transition $i$.
The exponent $\alpha$ determines how much prioritization is used, with $\alpha=0$ corresponding to the uniform case.

The first variant we consider is the direct, proportional prioritization where $p_i = |\delta_i| + \epsilon$, 
where $\epsilon$ is a small positive constant that prevents the edge-case of transitions not being revisited once their 
error is zero.
The second variant is an indirect, rank-based prioritization where $p_i = \frac{1}{\operatorname{rank}(i)}$,
where $\operatorname{rank}(i)$ is the rank of transition $i$ when the replay memory is sorted according to $|\delta_i|$.
In this case, $P$ becomes a power-law distribution with exponent $\alpha$.
Both distributions are monotonic in $|\delta|$, but the latter is likely to be more robust, 
as it is insensitive to outliers. 
Both variants of stochastic prioritization lead to large 
speed-ups over the uniform baseline on the Cliffwalk task, as shown on Figure~\ref{fig-baseline-oracle} (right).

{\bf Implementation}: To efficiently sample from distribution (\ref{eq-distr}), the complexity cannot depend on $N$.
For the rank-based variant, we can approximate the cumulative density function with a piecewise linear function with $k$ segments of equal probability.
The segment boundaries can be precomputed (they change only when $N$ or $\alpha$ change).
At runtime, we sample a segment, and then sample uniformly among the transitions
within it. This works particularly well in conjunction with a minibatch-based learning algorithm: 
choose $k$ to be the size of the minibatch, and sample exactly one transition from each segment --
this is a form of stratified sampling that has the added advantage of balancing out the minibatch 
(there will always be exactly one transition with high magnitude $\delta$, one with medium magnitude, etc).
The proportional variant is different, also admits an efficient implementation based
on a `sum-tree' data structure (where every node is the sum of its children, with the priorities as the leaf nodes), which can be efficiently updated and sampled from.  
See Appendix~\ref{app-atari-experiments-impl-details} for more additional details.

\subsection{Annealing the Bias}
\label{sec-anneal}

The estimation of the expected value with stochastic updates relies on those updates corresponding
to the same distribution as its expectation. Prioritized replay introduces bias because it changes this distribution in an uncontrolled fashion, and therefore changes the solution that the 
estimates will converge to (even if the policy and state distribution are fixed). 
We can correct this bias by using importance-sampling (IS) weights
\[
w_i = \left(\frac{1}{N} \cdot \frac{1}{P(i)}\right)^{\beta}
\]
that fully compensates for the non-uniform probabilities $P(i)$ if $\beta=1$.
These weights can be folded into the Q-learning update by using $w_i\delta_i$ instead of $\delta_i$ 
(this is thus \emph{weighted} IS, not ordinary IS, see e.g.\ \citealp{wis}).
For stability reasons, we always normalize weights by $1/\max_i w_i$ so that they only scale the update downwards.

In typical reinforcement learning scenarios, the unbiased nature of the updates is most important
near convergence at the end of training, as the process is highly non-stationary anyway, 
due to changing policies, state distributions and bootstrap targets;
we hypothesize that a small bias can be ignored in this context
(see also Figure~\ref{fig-atari-is} in the appendix for a case study of full IS correction on Atari).
We therefore exploit the flexibility of \emph{annealing} the amount of importance-sampling 
correction over time, by defining a schedule on the exponent $\beta$ that 
reaches $1$ only at the end of learning.
In practice, we linearly anneal $\beta$ from its initial value $\beta_0$ to $1$.
Note that the choice of this hyperparameter 
interacts with choice of prioritization exponent $\alpha$; increasing both simultaneously prioritizes sampling more aggressively at the same time as correcting for it more strongly.

Importance sampling has another benefit when combined with prioritized replay in the context of non-linear function approximation (e.g. deep neural networks):
here large steps can be very disruptive,
because the first-order approximation of the gradient is only reliable locally,
and have to be prevented with a smaller global step-size.
In our approach instead, prioritization makes sure high-error transitions are seen many times, while the IS correction reduces the gradient magnitudes 
(and thus the effective step size in parameter space),
and allowing the algorithm follow the curvature of highly non-linear optimization landscapes because the Taylor expansion is constantly re-approximated.

We combine our prioritized replay algorithm into a full-scale reinforcement learning agent, based on the state-of-the-art Double DQN algorithm. Our principal modification is to replace the uniform random sampling used by Double DQN with our stochastic prioritization and importance sampling methods (see Algorithm~\ref{alg-preplay}).

\begin{algorithm}[tb]
   \caption{Double DQN with proportional prioritization}
   \label{alg-preplay}
\begin{algorithmic}[1]
   \STATE {\bfseries Input:} minibatch $k$, step-size $\eta$, replay period $K$ and size $N$, exponents $\alpha$ and $\beta$, budget $T$.
   \STATE Initialize replay memory $\mathcal{H}=\emptyset$, $\Delta = 0$, $p_1=1$
   \STATE Observe $S_0$ and choose $A_0 \sim \pi_{\theta}(S_0)$
   \FOR{$t=1$ {\bfseries to} $T$}
   	  \STATE Observe $S_t, R_t, \gamma_t$
      \STATE Store transition $(S_{t-1}, A_{t-1}, R_t, \gamma_{t}, S_{t})$ in $\mathcal{H}$ with maximal priority $p_t = \max_{i<t} p_i$
      \IF{ $t \equiv 0 \mod K$ } 
	   	\FOR{$j=1$ {\bfseries to} $k$}
          \STATE Sample transition $j \sim P(j) = p_j^{\alpha} / \sum_i p_i^{\alpha}$
          \STATE Compute importance-sampling weight $w_j = \left(N\cdot P(j)\right)^{-\beta} / \max_i w_i$
          \STATE Compute TD-error $\delta_j =  R_j + \gamma_j Q_{\text{target}}\left(S_j, \arg\max_a Q(S_j, a)\right) - Q(S_{j-1}, A_{j-1}) $
          \STATE Update transition priority $p_j \leftarrow |\delta_j|$ 
          \STATE Accumulate weight-change $\Delta \leftarrow \Delta + w_j \cdot \delta_j \cdot \nabla_{\theta} Q(S_{j-1}, A_{j-1})$
        \ENDFOR
        \STATE Update weights $\theta \leftarrow \theta + \eta \cdot \Delta $, reset $\Delta = 0$
        \STATE From time to time copy weights into target network $\theta_{\text{target}} \leftarrow \theta$
      \ENDIF
      \STATE Choose action $A_t \sim \pi_{\theta}(S_t)$
   \ENDFOR
\end{algorithmic}
\end{algorithm}

\begin{figure}[t]
\vspace{-1em}
\centerline{
\includegraphics[width=1.05\textwidth]{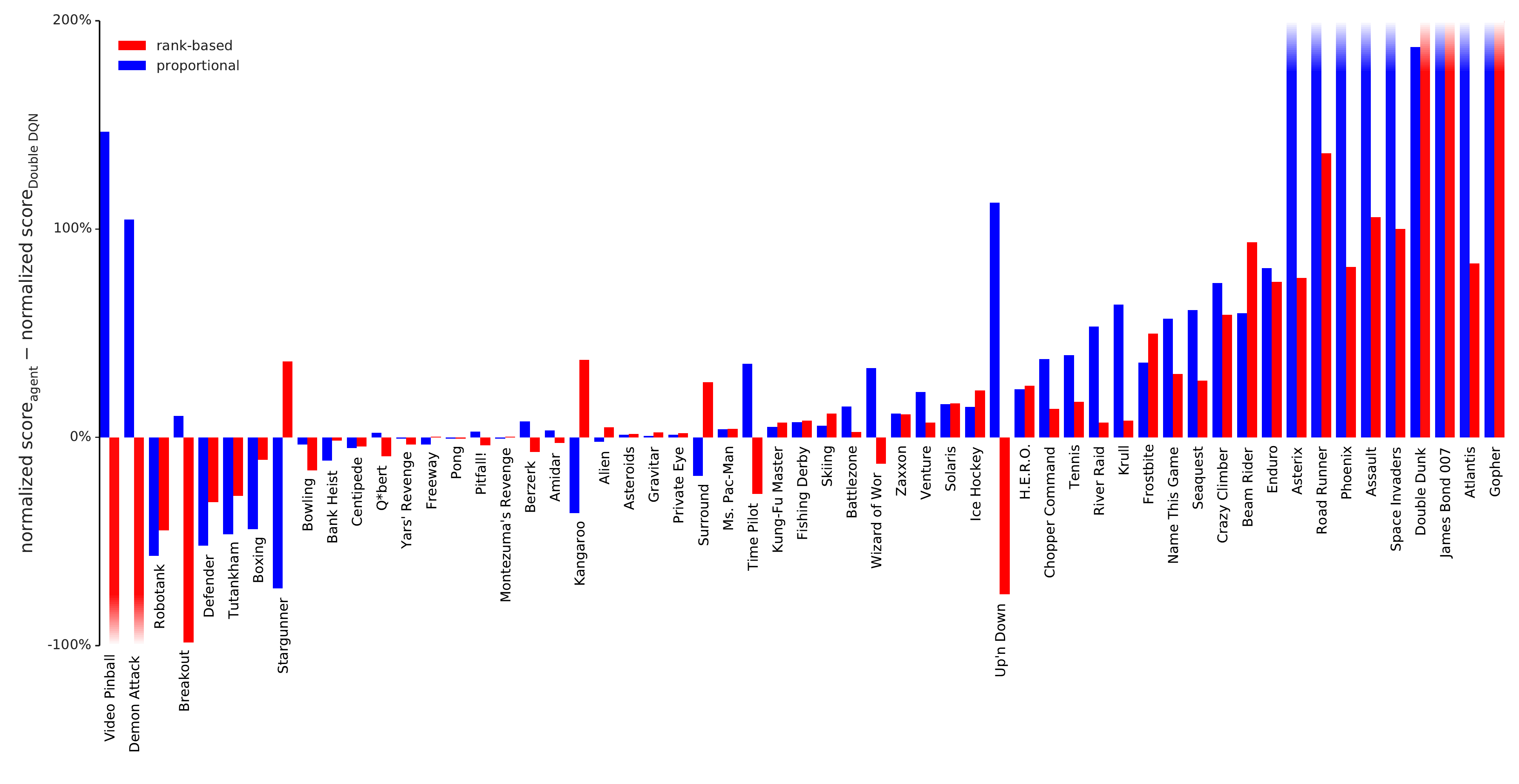}
}
\vspace{-1em}
\caption{
\label{fig-gen-bars}
Difference in normalized score (the gap between random and human is $100\%$) on 57 games with human starts, comparing Double
DQN with and without prioritized replay (rank-based variant in red, proportional in blue), showing substantial improvements in most games.  Exact scores are in Table~\ref{tab-atari-normalized-30human}.
See also Figure~\ref{fig-dqn-bars} where regular DQN is the baseline.
\vspace{-1em}
}
\end{figure}

\section{Atari Experiments} 

With all these concepts in place, 
we now investigate to what extent replay with such prioritized sampling 
can improve performance in realistic problem domains. 
For this, we chose the collection of Atari benchmarks~\citep{bellemare2012arcade} with their end-to-end RL from vision setup,
because they are popular and contain diverse sets of challenges, including delayed credit assignment, partial observability, and difficult function approximation~\citep{dqn-nature,double-dqn}.
Our hypothesis is that prioritized replay is generally useful, so that it will make learning with experience replay more efficient
without requiring careful problem-specific hyperparameter tuning.

We consider two baseline algorithms that use uniform experience replay, namely the version of the DQN algorithm from the Nature paper~\citep{dqn-nature},
and its recent extension Double DQN~\citep{double-dqn} that substantially improved the state-of-the-art by reducing the over-estimation bias with Double Q-learning~\citep{double-q}.  Throughout this paper we use the tuned version of the Double DQN algorithm.
For this paper, the most relevant component of these baselines is the replay mechanism:
all experienced transitions are stored in a sliding window memory that retains the
last $10^6$ transitions.
The algorithm processes minibatches of 32 transitions sampled uniformly from the memory. One minibatch update is done for each 4 new transitions entering the memory, so all experience is replayed 8 times on average. Rewards and TD-errors are clipped to fall within $[-1, 1]$ for stability reasons.

We use the identical neural network architecture, learning algorithm, replay memory and evaluation setup as for the baselines (see Appendix~\ref{app-atari-experiments}).
The only difference is the mechanism for sampling transitions from the replay memory, with is now done according to Algorithm~\ref{alg-preplay} instead of uniformly.
We compare the baselines to both variants of prioritized replay (rank-based and proportional).

Only a single hyperparameter adjustment was necessary compared to the baseline:
Given that prioritized replay picks high-error transitions more often, the typical gradient magnitudes
are larger, so we reduced the step-size $\eta$ by a factor 4 compared to the (Double) DQN setup.
For the $\alpha$ and $\beta_0$ hyperparameters that are introduced by prioritization, 
we did a coarse grid search (evaluated on a subset of 8 games), 
and found the sweet spot to be $\alpha = 0.7$, $\beta_0 = 0.5$ for the rank-based variant
and $\alpha = 0.6$, $\beta_0 = 0.4$ for the proportional variant.
These choices are trading off aggressiveness with robustness, but it is easy to revert to a behavior closer to the baseline by reducing $\alpha$ and/or increasing $\beta$.

\begin{figure}[tb]
\vspace{-1.5em}
\centerline{
\includegraphics[width=1\textwidth]{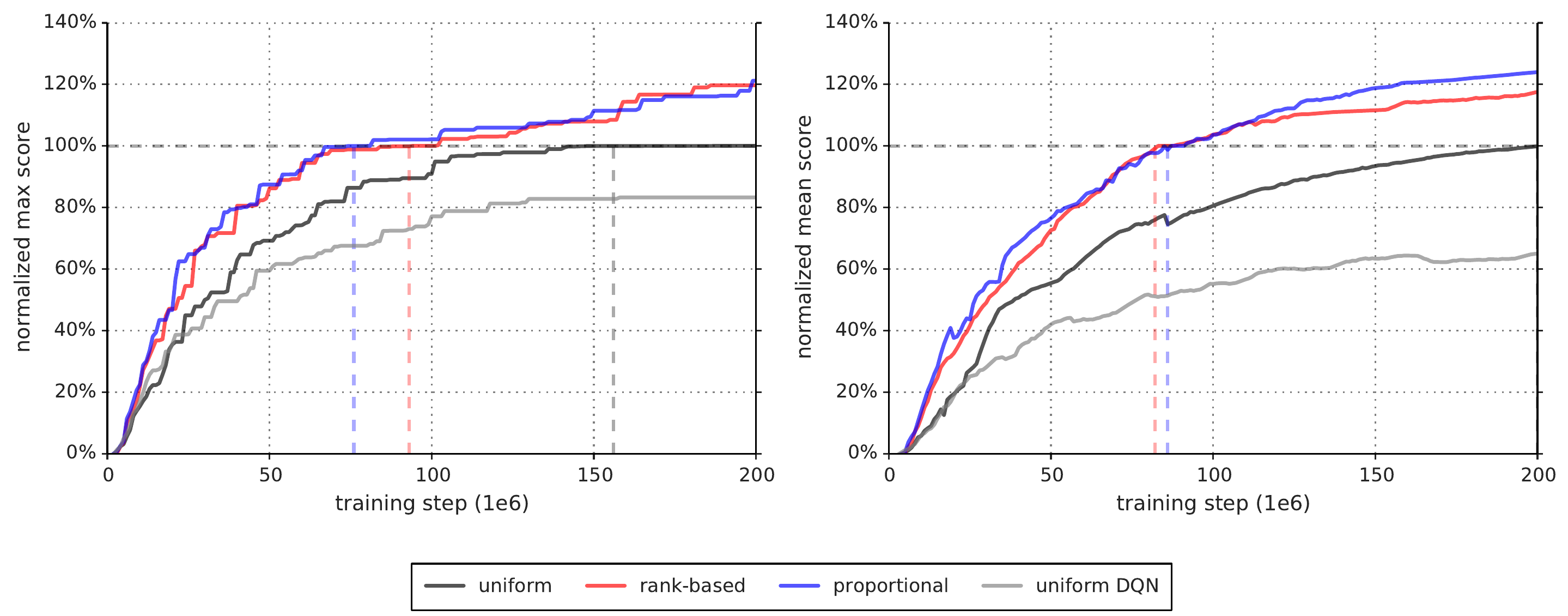}
}
\caption{
\label{fig-atari-progress}
Summary plots of learning speed. {\bf Left:} median over 57 games of the maximum baseline-normalized score achieved so far. The baseline-normalized score is calculated as in Equation~\ref{eq-norm-score} but using the maximum Double DQN score seen across training is used instead of the human score.  
The equivalence points are highlighted with dashed lines;
those are the steps at which the curves reach 100\%, (i.e., when the algorithm performs equivalently to Double DQN in terms of median over games). 
For rank-based and proportional prioritization these are at 47\% and 38\% of total training time.
{\bf Right:} Similar to the left, but using the mean instead of maximum, which captures cumulative performance rather than peak performance. 
Here rank-based and proportional prioritization reach the equivalence points at 41\% and 43\% of total training time, respectively.
For the detailed learning curves that these plots summarize, see Figure~\ref{fig-atari-all}.
\vspace{-1.em}
}
\end{figure}

We produce the results by running each variant with a single hyperparameter setting across all games, as was done for the baselines.
Our main evaluation metric is the \emph{quality of the best policy}, in terms of average score per episode, given start states sampled from human traces 
(as introduced in~\citealp{gorila} and used in~\citealp{double-dqn},
which requires more robustness and generalization as the agent cannot rely on repeating a single memorized trajectory).   
These results are summarized in Table~\ref{tab-atari-normalized-summary} and Figure~\ref{fig-gen-bars}, 
but full results and raw scores can be found in 
Tables~\ref{tab-atari-raw-30human} and~\ref{tab-atari-normalized-30human} in the Appendix.
A secondary metric is the \emph{learning speed}, which we summarize on Figure~\ref{fig-atari-progress}, with more detailed learning curves on Figures~\ref{fig-atari-all} and~\ref{fig-atari-subset-detailed}.

\begin{table*}[htb]
\centering
\makebox[\textwidth][c]{
\setlength{\tabcolsep}{4pt}
{


\begin{tabular}{|l|rr|rrr|}
\hline
& \multicolumn{2}{c}{DQN} & \multicolumn{3}{|c|}{Double DQN (tuned)} \\
\cline{2-6}
& baseline & rank-based & baseline & rank-based & proportional \\
\hline
\textbf{Median}  & 48\% & 106\% & 111\% & 113\% & 128\%\\
\textbf{Mean}  & 122\% & 355\% & 418\% & 454\% & 551\%\\
$\bm{>}$ \textbf{baseline}  & -- & 41 & -- & 38 & 42\\
$\bm{>}$ \textbf{human}  & 15 & 25 & 30 & 33 & 33\\
\textbf{\# games}  & 49 & 49 & 57 & 57 & 57\\
\hline
\end{tabular}


} 
} 
\vspace{-1em}
\caption{
\label{tab-atari-normalized-summary}
Summary of normalized scores. See Table~\ref{tab-atari-normalized-30human} in the appendix for full results.
}
\end{table*}

We find that adding prioritized replay to DQN leads to a substantial improvement in score on 41 out of 49 games
(compare columns 2 and 3 of Table~\ref{tab-atari-normalized-30human} or Figure~\ref{fig-dqn-bars} in the appendix), 
with the median normalized performance across 49 games increasing from $48\%$ to $106\%$. 
Furthermore, we find that the boost from prioritized experience replay is \emph{complementary} to the one 
from introducing Double Q-learning into DQN: 
performance increases another notch, leading to the current state-of-the-art on the Atari benchmark
(see Figure~\ref{fig-gen-bars}). 
Compared to Double DQN, the median performance across 57 games increased from $111\%$ to $128\%$, and the mean performance from $418\%$ to $551\%$ bringing
additional games such as River Raid, Seaquest and Surround to a human level for the first time, and making large jumps on others (e.g.\ Gopher, James Bond 007 or Space Invaders).
Note that mean performance is not a very reliable metric because a single game (Video Pinball) has a dominant contribution.
Prioritizing replay gives a performance boost on almost all games, 
and on aggregate, learning is twice as fast (see Figures~\ref{fig-atari-progress} and~\ref{fig-atari-subset-detailed}).
The learning curves on Figure~\ref{fig-atari-all} illustrate that while the two variants of prioritization usually lead to similar results,
there are games where one of them remains close to the Double DQN baseline while the other one leads to a big boost,
for example Double Dunk or Surround for the rank-based variant, and Alien, Asterix, Enduro, Phoenix or Space Invaders for the proportional variant.
Another observation from the learning curves is that compared to the uniform baseline, prioritization is effective at reducing the delay until performance gets off the ground in games that otherwise suffer from such a delay, such as Battlezone, Zaxxon or Frostbite. 

\section{Discussion}

In the head-to-head comparison between rank-based prioritization and proportional prioritization,
we expected the rank-based variant to be more robust because it is not affected by outliers nor error magnitudes. 
Furthermore, its heavy-tail property also guarantees that samples will be diverse,
and the stratified sampling from partitions of different errors 
will keep the total minibatch gradient at a stable magnitude throughout training.
On the other hand, the ranks make the algorithm blind 
to the relative error scales, which could incur a performance drop when there is structure in the distribution of
errors to be exploited, such as in sparse reward scenarios. Perhaps surprisingly, both variants perform similarly in practice; we suspect this is
due to the heavy use of clipping (of rewards and TD-errors) in the DQN algorithm, which removes outliers.
Monitoring the distribution of TD-errors as a function of time for a number of games
(see Figure~\ref{fig-transition-priorities} in the appendix), 
and found that it becomes close to a heavy-tailed distribution as learning progresses,
while still differing substantially across games;  this empirically validates the form of Equation~\ref{eq-distr}.
Figure~\ref{fig-priority-distribution}, in the appendix, shows how this distribution interacts with Equation~\ref{eq-distr} 
to produce the effective replay probabilities.

While doing this analysis, we stumbled upon another phenomenon (obvious in retrospect), 
namely that some fraction of the visited transitions are never replayed 
before they drop out of the sliding window memory, 
and many more are replayed for the first time only long after they are encountered.
Also, uniform sampling is implicitly 
biased toward out-of-date transitions that were generated by a policy that has typically seen hundreds of thousands of updates since.
Prioritized replay with its bonus for unseen transitions directly corrects the first of these issues, 
and also tends to help with the second one, as more recent transitions tend to have larger error -- 
this is because old transitions will have had more opportunities to have them corrected, and
because novel data tends to be less well predicted by the value function.

We hypothesize that deep neural networks interact with prioritized replay in another interesting way. 
When we distinguish learning the value given a representation (i.e., the top layers)
from learning an improved representation (i.e., the bottom layers),
then transitions for which the representation is good will quickly reduce their error and then be replayed much less,
increasing the learning focus on others where the representation is poor, 
thus putting more resources into distinguishing aliased states -- if the observations and network capacity allow for it.

\section{Extensions}

\begin{figure}
\vspace{-1.5em}
\centerline{
\includegraphics[width=0.5\textwidth]{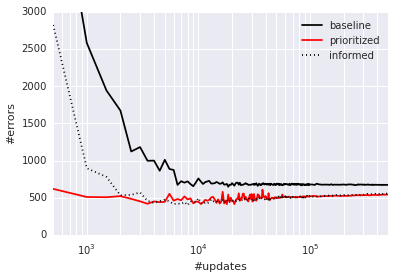}
\includegraphics[width=0.5\textwidth]{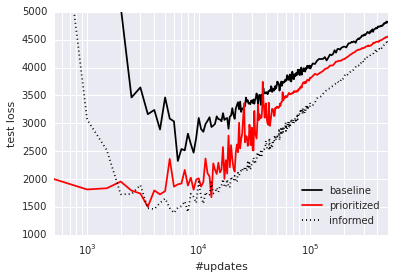}
}
\vspace{-1em}
\caption{
\label{fig-mnist}
Classification errors as a function of supervised learning updates on severely class-imbalanced MNIST. 
Prioritized sampling improves performance, leading to comparable errors on the test set,
and approaching the imbalance-informed performance (median of 3 random initializations). 
{\bf Left}: Number of misclassified test set samples. {\bf Right}: Test set loss, highlighting overfitting. 
\vspace{-1.5em}
}
\end{figure}

{\bf Prioritized Supervised Learning}:
The analogous approach to prioritized replay in the context of supervised learning 
is to sample non-uniformly from the dataset, each sample using a priority based on its last-seen error.
This can help focus the learning on those samples that
can still be learned from, devoting additional resources to the (hard) boundary cases, 
somewhat similarly to boosting~\citep{imbalance-ensemble}.
Furthermore, if the dataset is imbalanced, we hypothesize that samples from the rare classes 
will be sampled disproportionately often, because their errors shrink less fast,
and the chosen samples from the common classes will be those nearest to the decision boundaries,
leading to an effect similar to hard negative mining~\citep{hard-negative}.
To check whether these intuitions hold, we conducted a preliminary experiment 
on a class-imbalanced variant of the classical MNIST digit classification problem~\citep{mnist}, 
where we removed $99\%$ of the samples for digits $0,1,2,3,4$ in the training set,
while leaving the test/validation sets untouched (i.e., those retain class balance). 
We compare two scenarios: in the informed case, we reweight the errors of the impoverished classes artificially (by a factor $100$),
in the uninformed scenario, we provide no hint that the test distribution will  differ from the training distribution.
See Appendix~\ref{app-mnist-details} for the details of the convolutional neural network training setup.
Prioritized sampling (uninformed, with $\alpha=1$, $\beta=0$) outperforms the uninformed uniform baseline, 
and approaches the performance of the informed uniform baseline in terms of generalization (see Figure~\ref{fig-mnist});
again, prioritized training is also faster in terms of learning speed.

{\bf Off-policy Replay}: 
Two standard approaches to off-policy RL are rejection sampling and using importance sampling 
ratios $\rho$ to correct for how likely a transition would have been on-policy.
Our approach contains analogues to both these approaches, the replay probability $P$ and the IS-correction $w$.
It appears therefore natural to apply it to off-policy RL, if transitions are available in a replay memory.
In particular, we recover weighted IS with $w=\rho$, $\alpha=0$, $\beta=1$ and 
rejection sampling with $p=\min(1;\rho)$, $\alpha=1$, $\beta=0$, in the proportional variant.
Our experiments indicate that intermediate variants, possibly with annealing or ranking, could be more useful
in practice -- especially when IS ratios introduce high variance, i.e., when the policy of interest
differs substantially from the behavior policy in some states.
Of course, off-policy correction is complementary to our prioritization based on expected learning progress,
and the same framework can be used for a hybrid prioritization by defining $p = \rho \cdot |\delta|$, 
or some other sensible trade-off based on both $\rho$ and $\delta$.

{\bf Feedback for Exploration}: 
An interesting side-effect of prioritized replay is that the total number $M_i$ that a transition will end up being replayed varies widely, 
and this gives a rough indication of how useful it was to the agent.
This potentially valuable signal can be fed back to the exploration strategy that generates the transitions. 
For example, we could sample exploration hyperparameters 
(such as the fraction of random actions $\epsilon$, the Boltzmann temperature, or the amount of of intrinsic reward to mix in) 
from a parametrized distribution at the beginning of each episode, monitor the usefulness of the experience via $M_i$, and update 
the distribution toward generating more useful experience.
Or, in a parallel system like the Gorila agent~\citep{gorila},
it could guide resource allocation between a collection of concurrent but heterogeneous `actors', each with different exploration hyperparameters.

\label{sec-prio-memory}

{\bf Prioritized Memories}:
Considerations that help determine which transitions to replay are likely to also be relevant 
for determining which memories to store and when to erase them (e.g.\ when it becomes likely that
they will never be replayed anymore).
An explicit control over which memories to keep or erase 
can help reduce the required total memory size, because it reduces redundancy (frequently visited transitions will 
have low error, so many of them will be dropped), while automatically
adjusting for what has been learned already (dropping many of the `easy' transitions) 
and biasing the contents of the memory to where the errors remain high.
This is a non-trivial aspect, because memory requirements for DQN 
are currently dominated by the size of the replay memory, no longer by the size of the neural network.
Erasing is a more final decision than reducing the replay probability, 
thus an even stronger emphasis of diversity may be necessary,
for example by tracking the age of each transitions and using it to modulate the priority 
in such a way as to preserve sufficient old experience to prevent cycles~\citep[related to `hall of fame' ideas in multi-agent literature,][]{hall-of-fame}.
The priority mechanism is also flexible enough to permit integrating experience from
\emph{other sources}, such as from a planner or from human expert trajectories~\citep{dagger}, 
since knowing the source can be used to modulate each transition's priority, 
e.g.~in such a way as to preserve a sufficient fraction of external experience in memory.

\section{Conclusion}
This paper introduced prioritized replay, a method that can make
learning from experience replay more efficient.
We studied a couple of variants, devised implementations that scale to large 
replay memories, and found that prioritized replay speeds up learning by a factor 2
and leads to a new state-of-the-art of performance on the Atari benchmark.
We laid out further variants and extensions that hold promise, 
namely for class-imbalanced supervised learning.

\subsubsection*{Acknowledgments}
We thank our Deepmind colleagues, 
in particular Hado van Hasselt, Joseph Modayil, Nicolas Heess, Marc Bellemare, Razvan Pascanu, Dharshan Kumaran, Daan Wierstra, Arthur Guez, Charles Blundell, Alex Pritzel, Alex Graves, Balaji Lakshminarayanan, Ziyu Wang, Nando de Freitas, Remi Munos and Geoff Hinton for insightful discussions and feedback.

\bibliography{bib}
\bibliographystyle{iclr2016_conference}

\newpage

\appendix

\section{Prioritization Variants}
\label{sec-alternative-td}

The absolute TD-error is only one possible proxy for the ideal priority measure of expected learning progress.
While it captures the scale of potential improvement, it ignores inherent stochasticity in rewards or transitions, 
as well as possible limitations from partial observability or FA capacity; in other words, it is problematic
when there are unlearnable transitions. In that case, its \emph{derivative} -- which could be approximated by the difference between a transition's current $|\delta|$ and the $|\delta|$ when it was last replayed%
\footnote{Of course, more robust approximations would be a function of the history of all encountered $\delta$ values. 
In particular, one could imagine an RProp-style update~\citep{rprop} to priorities that increase the priority while 
the signs match, and reduce it whenever consecutive errors (for the same transition) differ in sign.}
-- may be more useful. 
This measure is less immediately available, however, and is influenced by whatever was replayed in the meanwhile, which increases its variance. 
In preliminary experiments, we found it  did not outperform $|\delta|$, but this may 
say more about the class of (near-deterministic) environments we investigated, than about the measure itself.

An orthogonal variant is to consider the norm of the \emph{weight-change} induced by replaying a transition -- this can be effective if
the underlying optimizer employs adaptive step-sizes that reduce gradients in high-noise directions~\citep{pesky,adam}, 
thus placing the burden of distinguishing between learnable and unlearnable transitions on the optimizer.

It is possible to modulate prioritization by not treating positive TD-errors the same than negative ones; we can for example invoke the
Anna Karenina principle~\citep{jdiamond-zebras}, interpreted to mean that there are many ways in which 
a transition can be less good than expected, but only one in which can be better, to introduce an \emph{asymmetry} and prioritize
positive TD-errors more than negative ones of equal magnitude, because the former are more likely to be informative.
Such an asymmetry in replay frequency was also observed in rat studies~\citep{dharsh4}.
Again, our preliminary experiments with such variants were inconclusive.

The evidence from neuroscience suggest that a prioritization based on episodic return rather than expected learning progress may be useful too~\cite{dharsh1,dharsh2,dharsh3}. For this case, we could boost the replay probabilities of entire episodes, instead of transitions, or boost individual transitions by their observed return-to-go (or even their value estimates).

For the issue of preserving sufficient diversity (to prevent overfitting, premature convergence or impoverished representations), 
there are alternative solutions 
to our choice of introducing stochasticity, for example, the priorities could be modulated
by a novelty measure in observation space. 
Nothing prevents a hybrid approach where some fraction of the elements (of each minibatch) are sampled according to one priority measure and the rest according to another one, introducing additional diversity.
An orthogonal idea is to increase priorities of transitions that have not been 
replayed for a while, by introducing an explicit \emph{staleness bonus} that guarantees that every transition 
is revisited from time to time, with that chance increasing at the same rate as its last-seen
TD-error becomes stale.
In the simple case where this bonus grows linearly with time, 
this can be implemented at no additional cost by subtracting a quantity proportional to the global step-count from the new priority on any update.%
\footnote{
If bootstrapping is used with policy iteration, such that the target values come from separate network (as is the case for DQN), 
then there is a large increase in uncertainty about the priorities when the target network is updated in the outer iteration.
At these points, the staleness bonus is increased in proportion to the number of individual (low-level) updates
that happened in-between.
}

In the particular case of RL with bootstrapping from value functions, it is possible to exploit the sequential structure 
of the replay memory using the following intuition:
a transition that led to a large amount of learning (about its outgoing state) 
has the potential to change the bootstrapping target for all transitions leading into that state, and
thus there is more to be learned about these. 
Of course we know at least one of these, namely the historic \emph{predecessor} transition,
and so boosting its priority makes it more likely to be revisited soon.
Similarly to eligibility traces, this lets information trickle backward from a future outcome to
the value estimates of the actions and states that led to it.
In practice, we add $|\delta|$ of the current transition to predecessor transition's priority, but only if the predecessor transition
is not a terminal one. This idea is related to `reverse replay' observed in rodents~\cite{dharsh3}, and to a recent extension of prioritized sweeping~\citep{harm}.

\section{Experimental Details}

\subsection{Blind Cliffwalk}
\label{app-microzuma}

For the Blind Cliffwalk experiments (Section~\ref{sec:example} and following), we 
use a straight-forward Q-learning~\citep{q-learning} setup.
The Q-values are represented using either a tabular look-up table,
or a linear function approximator, in both cases represented $Q(s,a) := \theta^{\top} \phi(s, a)$.
For each transition, we compute its TD-error using:
\begin{eqnarray}
\label{eq-td-error}
\delta_t & := & R_t + \gamma_t \max_a Q(S_{t}, a) - Q(S_{t-1}, A_{t-1}) 
\end{eqnarray}
and update the parameters using stochastic gradient ascent:
\begin{eqnarray}
\label{eq-qlearning}
\theta & \leftarrow & \theta + \eta \cdot \delta_t \cdot \nabla_{\theta} Q\bigr|_{S_{t-1}, A_{t-1}} = \theta + \eta \cdot \delta_t \cdot \phi(S_{t-1}, A_{t-1})
\end{eqnarray}
For the linear FA case we use a very simple encoding of state as a 1-hot vector (as for tabular), but concatenated with a constant bias feature of value 1.
To make generalization across actions impossible, we alternate which action is `right' and which one is `wrong' for each state.
All elements are initialized with small values near zero, $\theta_i \sim \mathcal{N}(0, 0.1)$.

We vary the size of the problem (number of states $n$) from 2 to 16. 
The discount factor is set to $\gamma=1-\frac{1}{n}$ which keeps values on approximately the same scale independently of $n$.
This allows us to used a fixed step-size of $\eta=\frac{1}{4}$ in all experiments.

The replay memory is filled by exhaustively executing all $2^n$ possible sequences of actions until termination (in random order). 
This guarantees that exactly one sequence will succeed and hit the final reward, and all others will fail with zero reward.
The replay memory contains all the relevant experience (the total number of transitions is $2^{n+1}-2$), 
at the frequency that it would be encountered when acting online with a random behavior policy.
Given this, we can in principle learn until convergence by just increasing the amount of computation;
here, convergence is defined as a mean-squared error (MSE) between the Q-value estimates and the ground truth below $10^{-3}$.

\subsection{Atari Experiments}
\label{app-atari-experiments}

\subsubsection{Implementation Details}
\label{app-atari-experiments-impl-details}

Prioritizing using a replay memory with $N=10^6$ transitions introduced some performance challenges.  Here we describe a number of things we did to minimize additional run-time and memory overhead, extending the discussion in Section~\ref{sec-preplay}.

\paragraph{Rank-based prioritization} Early experiments with Atari showed that maintaining a sorted data structure of $10^6$ transitions with constantly changing TD-errors dominated running time.  Our final solution  was to store transitions in a priority queue implemented with an array-based binary heap.  The heap array was then directly used as an approximation of a sorted array, which is infrequently sorted once every $10^6$ steps to prevent the heap becoming too unbalanced.  This is an unconventional use of a binary heap, however our tests on smaller environments showed learning was unaffected compared to using a perfectly sorted array.  This is likely due to the last-seen TD-error only being a proxy for the usefulness of a transition and our use of stochastic prioritized sampling.  A small improvement in running time came from avoiding excessive recalculation of partitions for the sampling distribution.  We reused the same partition for values of $N$ that are close together and by updating $\alpha$ and $\beta$ infrequently.  Our final implementation for rank-based prioritization produced an additional 2\%-4\% increase in running time and negligible additional memory usage.  This could be reduced further in a number of ways, e.g.\ with a more efficient heap implementation, but it was good enough for our experiments.

\paragraph{Proportional prioritization}
The `sum-tree' data structure used here is very similar in spirit to the array representation of a binary heap.  However, instead of the usual heap property, the value of a parent node is the sum of its children.  Leaf nodes store the transition priorities and the internal nodes are intermediate sums, with the parent node containing the sum over all priorities, $p_{\text{total}}$.  This provides a efficient way of calculating the cumulative sum of priorities, allowing $O(\log N)$ updates and sampling.  To sample a minibatch of size $k$, the range $[0, p_{\text{total}}]$ is divided equally into $k$ ranges.  Next, a value is uniformly sampled from each range.  Finally the transitions that correspond to each of these sampled values are retrieved from the tree.  Overhead is similar to rank-based prioritization.

As mentioned in Section~\ref{sec-anneal}, whenever importance sampling is used, all weights $w_i$ were scaled so that $\max_i w_i = 1$.  We found that this worked better in practice as it kept all weights within a reasonable range, avoiding the possibility of extremely large updates. It is worth mentioning that this normalization interacts with annealing on $\beta$: as $\beta$ approaches 1, the normalization constant grows, which reduces the effective average update in a similar way to annealing the step-size $\eta$.

\subsubsection{Hyperparameters}

Throughout this paper our baseline was DQN and the tuned version of Double DQN.  We tuned hyperparameters over a subset of Atari games: Breakout, Pong, Ms.\ Pac-Man, Q*bert, Alien, Battlezone, Asterix.  Table~\ref{tab-atari-hyperparameters} lists the values tried and Table~\ref{tab-atari-best-agent-hyperparameters} lists the chosen parameters.

\begin{table*}[!hp]
\centering

\makebox[\textwidth][c]{
\setlength{\tabcolsep}{4pt}
{\renewcommand{\arraystretch}{1}

\begin{tabular}{|c|l|}
\hline
Hyperparameter & Range of values\\
\hline
$\alpha$ & 0, 0.4, 0.5, 0.6, 0.7, 0.8 \\
$\beta$  & 0, 0.4, 0.5, 0.6, 1 \\
$\eta$   & $\eta_{\text{baseline}}, \eta_{\text{baseline}} / 2, \eta_{\text{baseline}} / 4, \eta_{\text{baseline}} / 8$ \\
\hline
\end{tabular}

} 
} 

\caption{
\label{tab-atari-hyperparameters}
Hyperparameters considered in experiments.  Here $\eta_{\text{baseline}} = 0.00025$.
}
\end{table*}

\begin{table*}[!hp]
\centering

\makebox[\textwidth][c]{
\setlength{\tabcolsep}{4pt}
{\renewcommand{\arraystretch}{1}

\begin{tabular}{|l|rr|rrr|}
\hline
& \multicolumn{2}{c}{DQN} & \multicolumn{3}{|c|}{Double DQN (tuned)} \\
\hline
Hyperparameter & baseline & rank-based & baseline & rank-based & proportional \\
\hline
$\alpha$ (priority) & 0       & $0.5 \rightarrow 0$         & 0       & 0.7                           & 0.6 \\
$\beta$ (IS) & 0       & 0                           & 0       & $0.5 \rightarrow 1$           & $0.4 \rightarrow 1$ \\
$\eta$ (step-size)  & 0.00025 & $\eta_{\text{baseline}} / 4$ & 0.00025 & $\eta_{\text{baseline}} / 4$ & $\eta_{\text{baseline}} / 4$ \\

\hline
\end{tabular}

} 
} 

\caption{
\label{tab-atari-best-agent-hyperparameters}
Chosen hyperparameters for prioritized variants of DQN.  Arrows indicate linear annealing, where the limiting value is reached at the end of training.  Note the rank-based variant with DQN as the baseline is an early version without IS.  Here, the bias introduced by prioritized replay was instead corrected by annealing $\alpha$ to zero.
}
\end{table*}

\subsubsection{Evaluation}
\label{sec-evaluation}

We evaluate our agents using the \emph{human starts} evaluation method described in \citep{double-dqn}.  Human starts evaluation uses start states sampled randomly from human traces.  The \emph{test} evaluation that agents periodically undergo during training uses start states that are randomized by doing a random number of no-ops at the beginning of each episode.  Human starts evaluation averages the score over 100 evaluations of 30 minutes of game time.  All learning curve plots show scores under the test evaluation and were generated using the same code base, with the same random seed initializations.

Table~\ref{tab-atari-evaluation-method} and Table~\ref{tab-atari-epsilon} show evaluation method differences and the $\epsilon$ used in the $\epsilon$-greedy policy for each agent during evaluation.  The agent evaluated with the human starts evaluation is the best agent found during training as in \citep{double-dqn}.

\begin{table*}[!hpt]
\centering

\makebox[\textwidth][c]{
\setlength{\tabcolsep}{4pt}
{\renewcommand{\arraystretch}{1}

\begin{tabular}{|l|rrrl|}
\hline
Evaluation method & Frames & Emulator time & Number of evaluations & Agent start point \\
\hline
Human starts & 108,000 & 30 mins & 100 & human starts\\
Test & 500,000 & 139 mins & 1 & up to 30 random no-ops\\
\hline
\end{tabular}

} 
} 

\caption{
\label{tab-atari-evaluation-method}
Evaluation method comparison.
}
\end{table*}

\begin{table*}[!hpt]
\centering

\makebox[\textwidth][c]{
\setlength{\tabcolsep}{4pt}
{\renewcommand{\arraystretch}{1}

\begin{tabular}{|l|rr|rrr|}

\hline
 & \multicolumn{2}{c}{DQN} & \multicolumn{3}{|c|}{Double DQN (tuned)} \\
\hline
Evaluation method & baseline & rank-based & baseline & rank-based & proportional\\
\hline
Human starts & 0.05 & 0.01 & 0.001 & 0.001 & 0.001 \\
Test          & 0.05 & 0.05 & 0.01 &  0.01 &  0.01 \\
\hline
\end{tabular}

} 
} 

\caption{
\label{tab-atari-epsilon}
The $\epsilon$ used in the $\epsilon$-greedy policy for each agent, for each evaluation method.
}
\end{table*}

Normalized score is calculated as in \citep{double-dqn}:
\begin{equation}
\label{eq-norm-score}
	\atariscore{normalized} = \frac{\atariscore{agent} - \atariscore{random}}{\left|\atariscore{human} - \atariscore{random}\right|}
\end{equation}

Note the absolute value of the denominator is taken.  This only affects Video Pinball where the random score is higher than the human score.  Combined with a high agent score, Video Pinball has a large effect on the mean normalized score.  We continue to use this so our normalized scores are comparable.

\subsection{Class-imbalanced MNIST}
\label{app-mnist-details}
\subsubsection{Dataset Setup}
In our supervised learning setting we modified MNIST to obtain a new training dataset with a significant label imbalance. This new dataset was obtained by considering a small subset of the samples that correspond to the first 5 digits (0, 1, 2, 3, 4) and all of the samples that correspond to the remaining 5 labels (5, 6, 7, 8, 9). For each of the first 5 digits we randomly sampled 1\% of the available examples, i.e., 1\% of the available 0s, 1\% of the available 1s etc. In the resulting dataset there are examples of all 10 different classes but it is highly imbalanced since there are 100 times more examples that correspond to the 5, 6, 7, 8, 9 classes than to the 0, 1, 2, 3, 4 ones. In all our experiments we used the original MNIST \emph{test} dataset without removing any samples. 

\begin{figure}[!h]
\centerline{
\includegraphics[width=1.\textwidth,trim={0 10cm 12cm 3cm},clip]{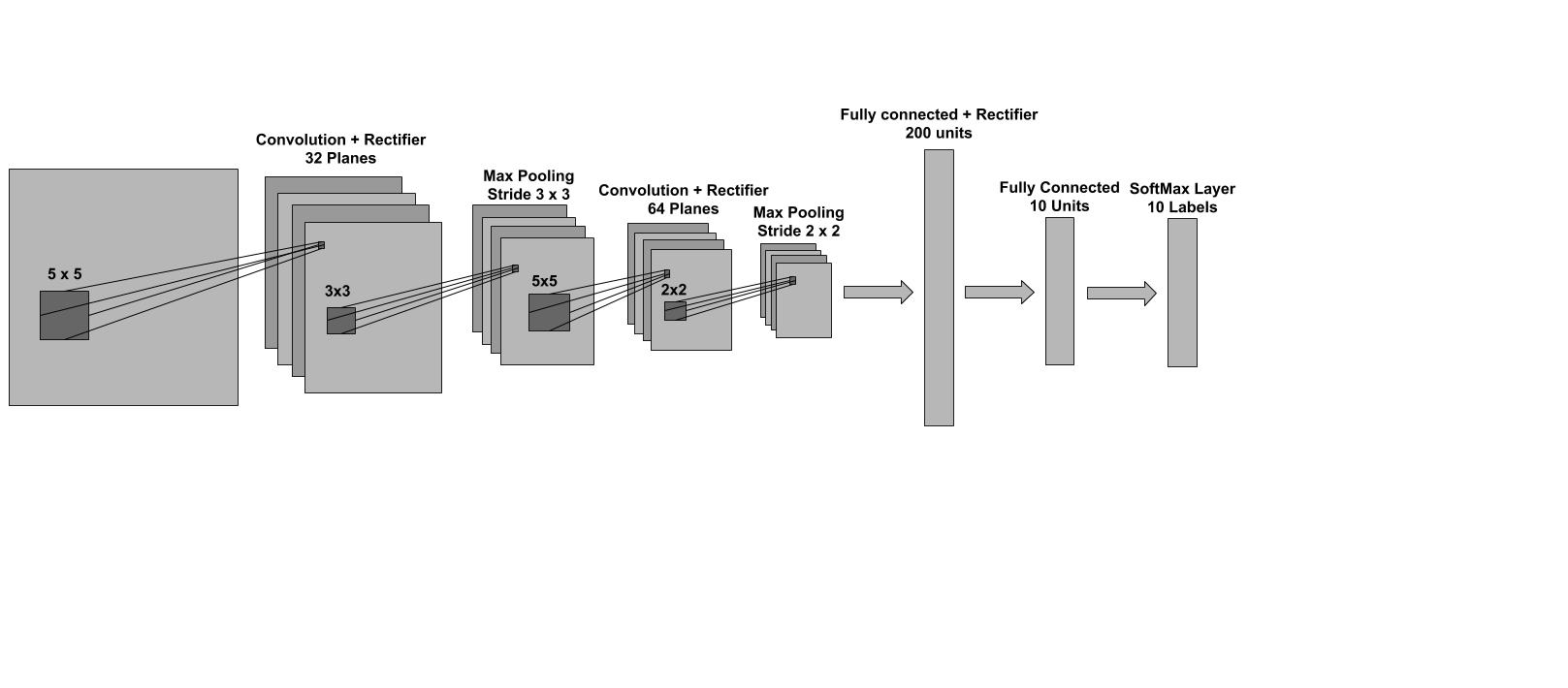}
}
\caption{
\label{convolutional-network}
}
The architecture of the feed-forward network used in the Prioritized Supervised Learning experiments.
\end{figure}

\subsubsection{Training Setup}
In our experiments we used a 4 layer feed-forward neural network with an architecture similar to that of LeNet5~\citep{lenet}. This is a 2 layer convolutional neural network followed by 2 fully connected layers at the top. Each convolutional layer is comprised of a pure convolution followed by a rectifier non-linearity and a subsampling max pooling operation. The two fully-connected layers in the network are also separated by a rectifier non-linearity. The last layer is a softmax which is used to obtain a normalized distribution over the possible labels. The complete architecture is shown in Figure~\ref{convolutional-network}, and is implemented using Torch7~\citep{torch7}.
The model was trained using stochastic gradient descent with no momentum and a minibatch of size 60. In all our experiments we considered 6 different step-sizes (0.3, 0.1, 0.03, 0.01, 0.003 and 0.001) and for each case presented in this work, we selected the step-size that lead to the best (balanced) validation performance. We used the negative log-likelihood loss criterion and we ran experiments with both the weighted and unweighted version of the loss. In the weighted case the loss of the examples that correspond to the first 5 digits (0, 1, 2, 3, 4) was scaled by a factor of a 100 to accommodate the label imbalance in the training set described above.

\begin{figure}[p]
\centerline{
\includegraphics[width=1.1\textwidth]{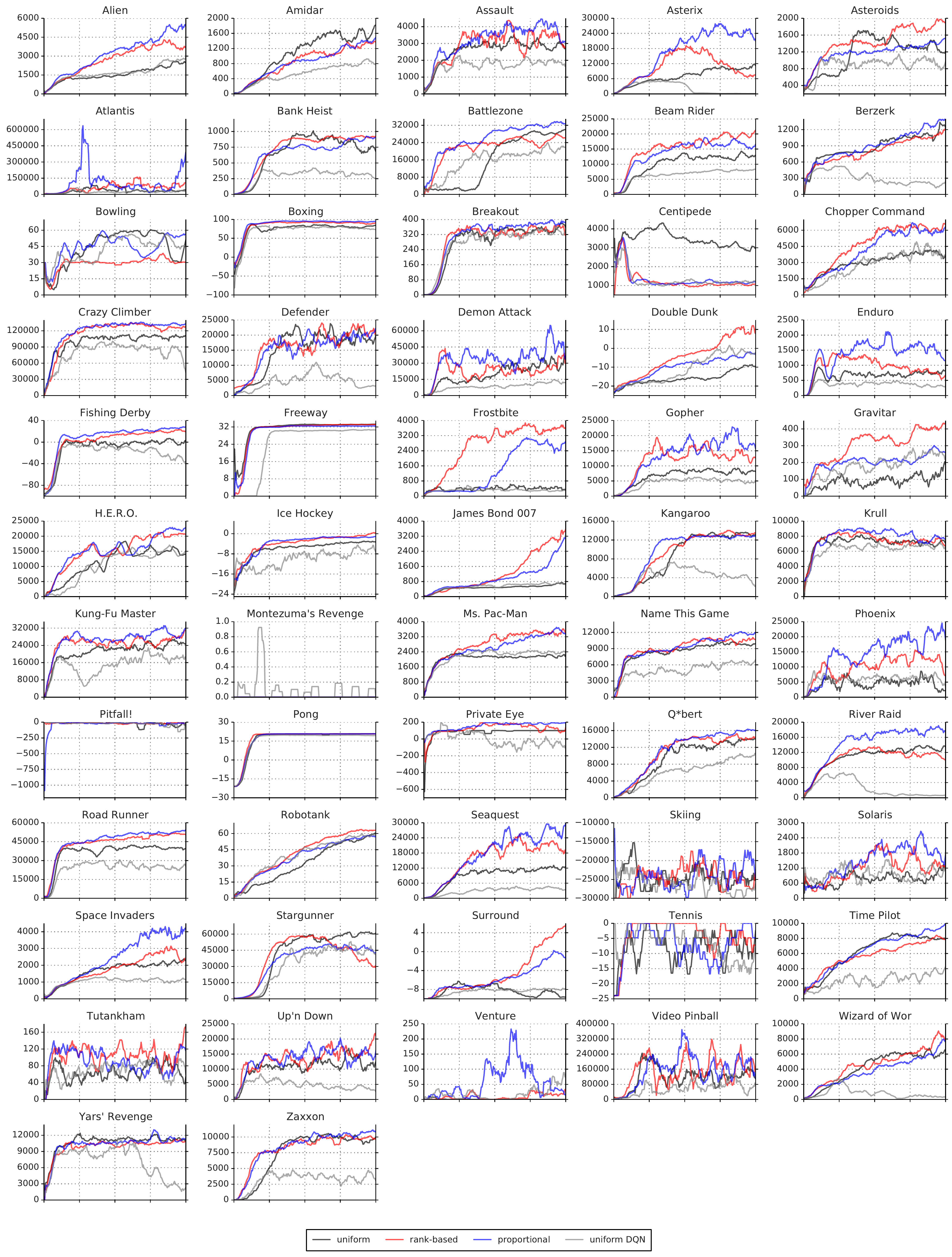}
}
\caption{
\label{fig-atari-all}
Learning curves (in raw score) for Double DQN (uniform baseline, in black), with rank-based prioritized replay (red), proportional prioritization (blue), for all 57 games of the Atari benchmark suite. Each curve corresponds to a single training run over 200 million unique frames, using test evaluation (see Section~\ref{sec-evaluation}),  with a moving average smoothed over 10 points. Learning curves for the original DQN are in gray. See Figure~\ref{fig-atari-subset-detailed} for a more detailed view on a subset of these. 
}
\end{figure}

\begin{figure}[p]
\centerline{
\includegraphics[width=1.1\textwidth]{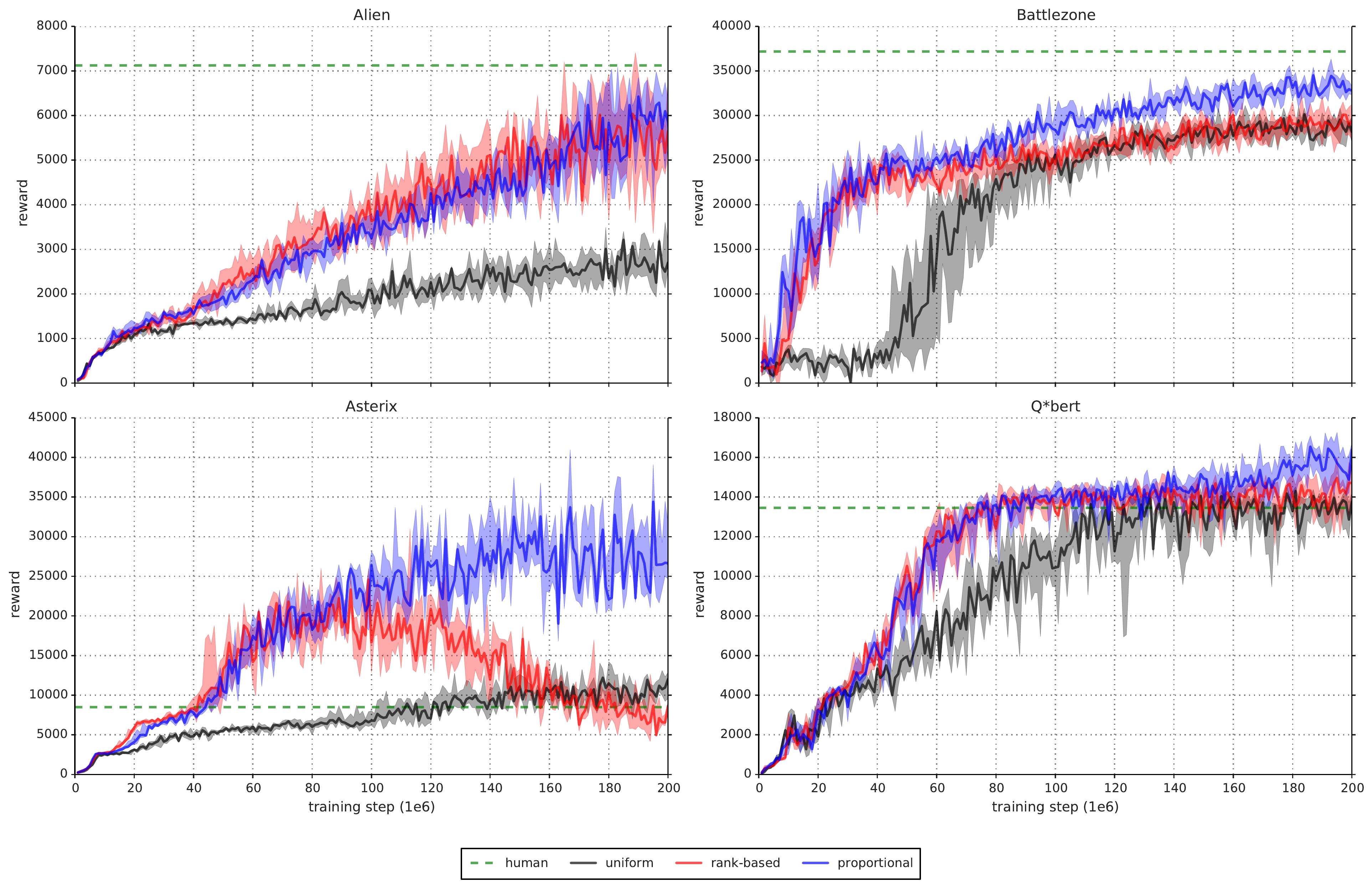}
}
\caption{
\label{fig-atari-subset-detailed}
Detailed learning curves for rank-based (red) and proportional (blue) prioritization, as compared to the uniform Double DQN baseline (black) on a selection of games.
The solid lines are the median scores, and the shaded area denotes the interquartile range across 8 random initializations.  The dashed green lines are human scores.
While the variability between runs is substantial, there are significant differences in final achieved score, and also in learning speed.
}
\end{figure}

\begin{figure}[p]
\centerline{
\includegraphics[width=0.9\textwidth]{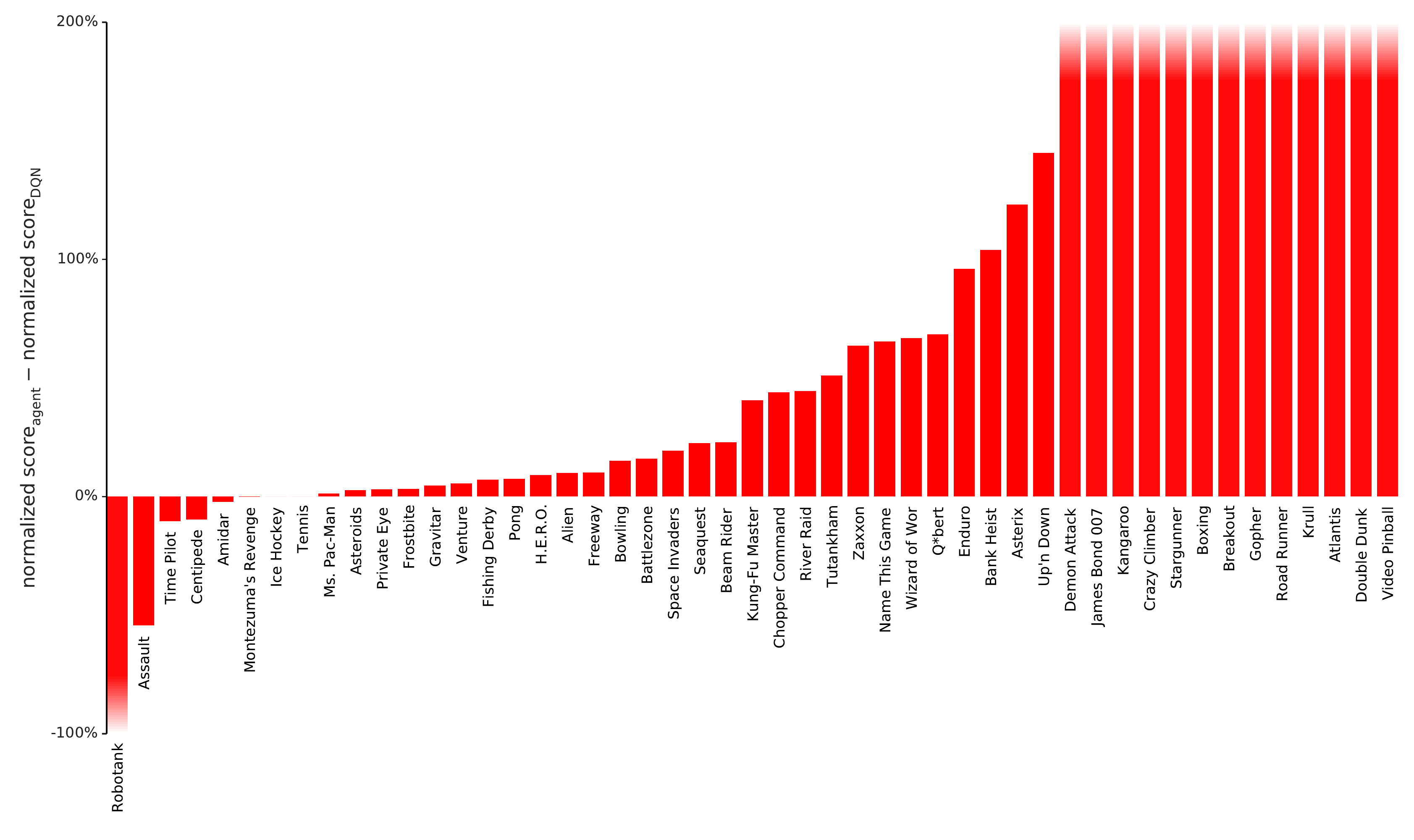} 
}
\caption{
\label{fig-dqn-bars}
Difference in normalized score (the gap between random and human is $100\%$) on 49 games with human starts, comparing DQN with and without rank-based prioritized replay, 
showing substantial improvements in many games.  Exact scores are in Table~\ref{tab-atari-normalized-30human}.
See also Figure~\ref{fig-gen-bars} where Double DQN is the baseline.
}
\end{figure}

\begin{figure}[p]
\centerline{
\includegraphics[width=1.1\textwidth]{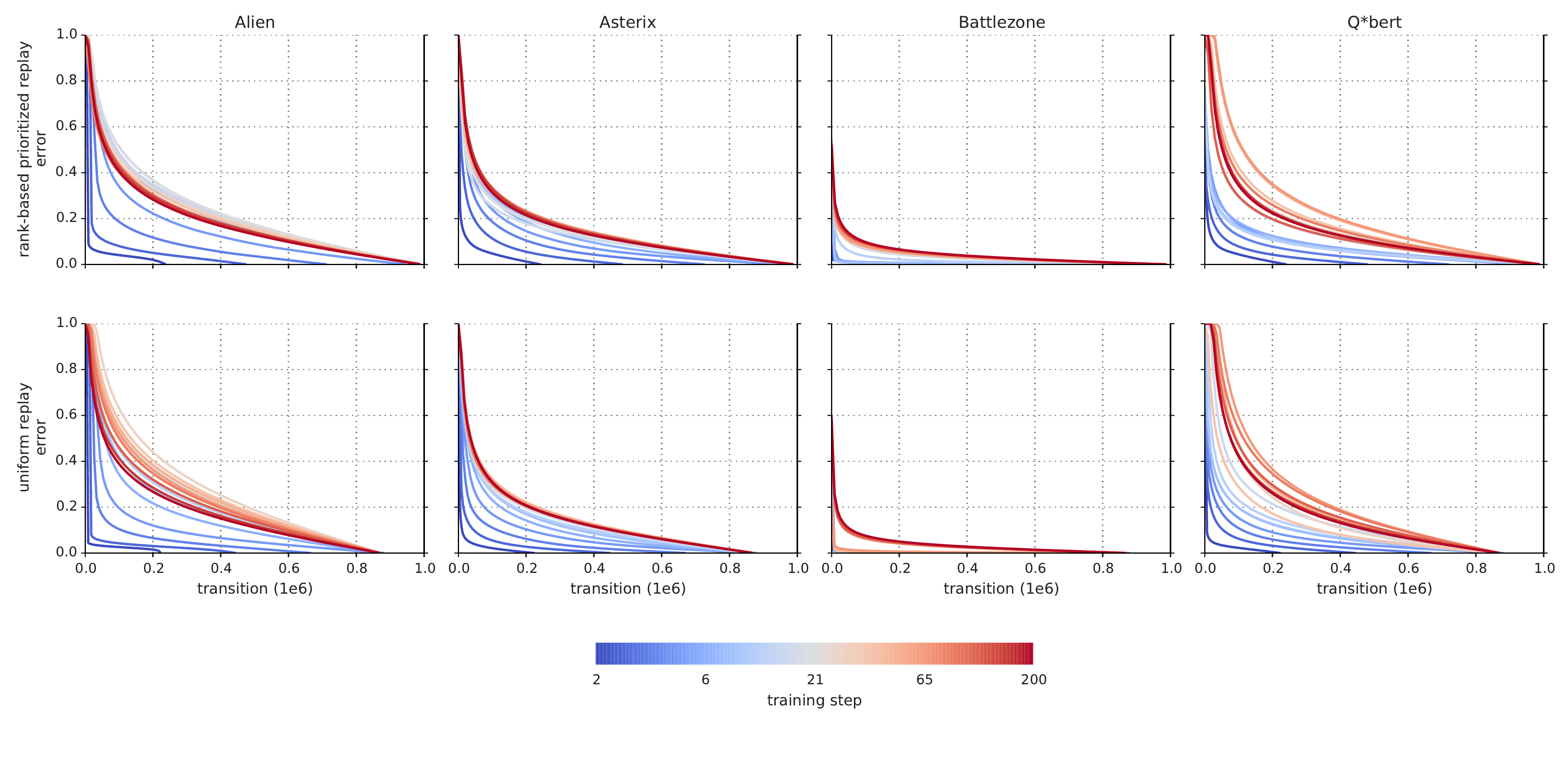}
}
\caption{
\label{fig-transition-priorities}
Visualization of the last-seen absolute TD-error for all transitions in the replay memory, sorted, for a selection of Atari games. 
The lines are color-coded by the time during learning, at a resolution of $10^6$ frames, with the coldest colors in the beginning and the warmest toward the end of training.
We observe that in some games it starts quite peaked but quickly becomes spread out, following approximately a heavy-tailed distribution. This phenomenon happens for both rank-based prioritized replay (top) and uniform replay (bottom) but is faster for prioritized replay.
}
\end{figure}

\begin{figure}[p]
\centerline{
\includegraphics[width=1.1\textwidth]{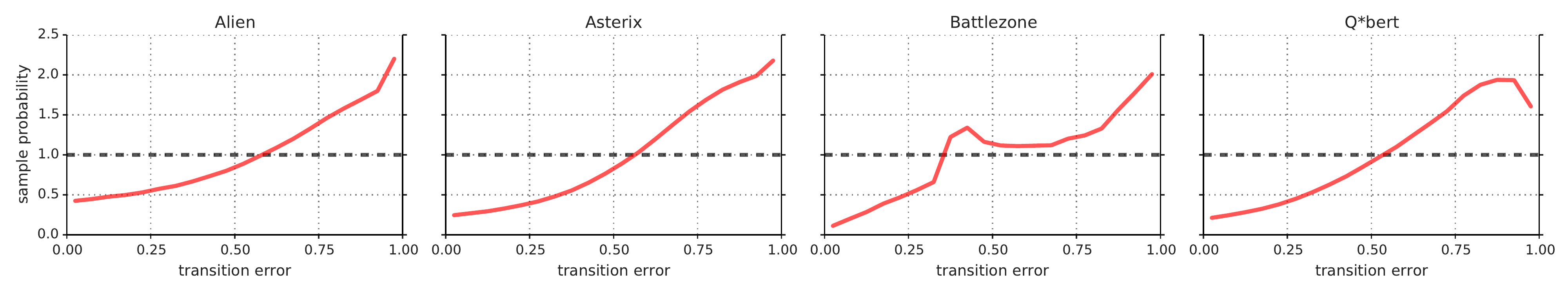}
}
\caption{
\label{fig-priority-distribution}
{\bf Effective replay probability}, as a function of absolute TD-error, for the rank-based prioritized replay variant near the start of training. 
This shows the effect of Equation~\ref{eq-distr} with $\alpha=0.7$ in practice, compared to the uniform baseline (dashed horizontal line).
The effect is irregular, but qualitatively similar for a selection of games.
}
\end{figure}

\begin{figure}[p]
\centerline{
\includegraphics[width=1.1\textwidth]{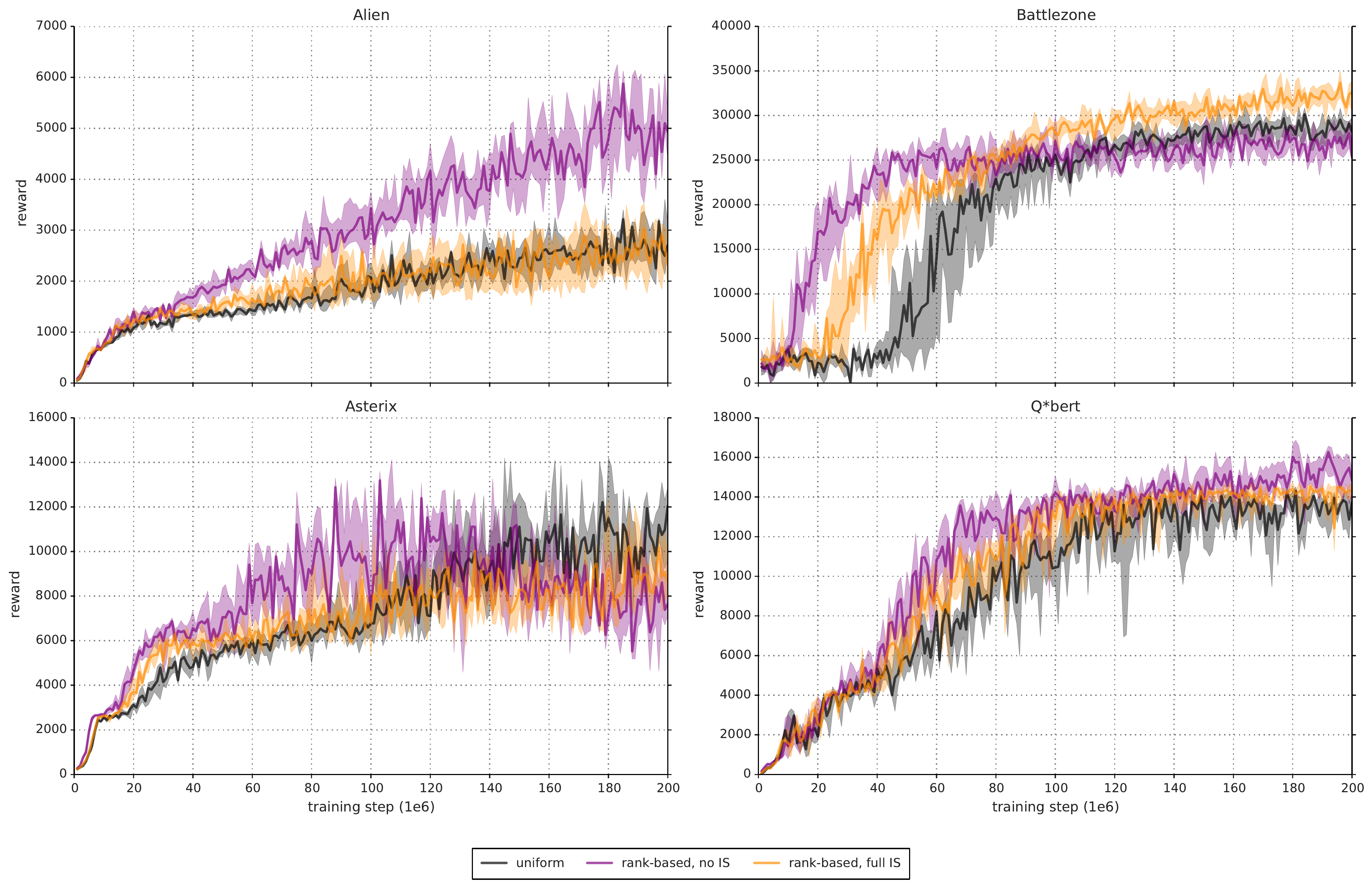}
}
\caption{
\label{fig-atari-is}
{\bf Effect of importance sampling}: These learning curves (as in Figure~\ref{fig-atari-subset-detailed})
show how rank-based prioritization is affected by full importance-sampling correction (i.e., $\beta=1$, in orange),
as compared to the uniform baseline (black, $\alpha=0$) and pure, uncorrected prioritized replay (violet, $\beta=0$),
on a few selected games. The shaded area corresponds to the interquartile range.  The step-size for full IS correction is the same as for uniform replay. For uncorrected prioritized replay, the step-size is reduced by a factor of 4.
Compared to uncorrected prioritized replay, importance sampling makes learning less aggressive, 
leading on the one hand to slower initial learning, 
but on the other hand to a smaller risk of premature convergence and sometimes better ultimate results.
Compared to uniform replay, fully corrected prioritization is on average better.
}
\end{figure}

\begin{table*}[p]
\centering

\makebox[\textwidth][c]{
\setlength{\tabcolsep}{1.25pt}
{\renewcommand{\arraystretch}{0.8}


\begin{tabular}{|l|rr|rrr|}
\hline
& \multicolumn{2}{c}{DQN} & \multicolumn{3}{|c|}{Double DQN (tuned)} \\
\hline
Game & baseline & rank-based & baseline & rank-based & proportional \\
\hline
Alien & 7\% & 17\% & 14\% & \textbf{19\%} & 12\%\\
Amidar & 8\% & 6\% & 10\% & 8\% & \textbf{14\%}\\
Assault & 685\% & 631\% & 1276\% & 1381\% & \textbf{1641\%}\\
Asterix & -1\% & 123\% & 226\% & 303\% & \textbf{431\%}\\
Asteroids & -0\% & 2\% & 1\% & \textbf{2\%} & 2\%\\
Atlantis & 478\% & 1480\% & 2335\% & 2419\% & \textbf{4425\%}\\
Bank Heist & 25\% & 129\% & \textbf{139\%} & 137\% & 128\%\\
Battlezone & 48\% & 63\% & 72\% & 75\% & \textbf{87\%}\\
Beam Rider & 57\% & 80\% & 117\% & \textbf{210\%} & 176\%\\
Berzerk &  & 22\% & 40\% & 33\% & \textbf{47\%}\\
Bowling & 5\% & 20\% & \textbf{31\%} & 15\% & 27\%\\
Boxing & 246\% & 641\% & \textbf{676\%} & 665\% & 632\%\\
Breakout & 1149\% & \textbf{1823\%} & 1397\% & 1298\% & 1407\%\\
Centipede & 22\% & 12\% & \textbf{23\%} & 19\% & 18\%\\
Chopper Command & 29\% & \textbf{73\%} & 34\% & 48\% & 72\%\\
Crazy Climber & 179\% & 429\% & 448\% & 507\% & \textbf{522\%}\\
Defender &  & 151\% & \textbf{207\%} & 176\% & 155\%\\
Demon Attack & 390\% & 596\% & 2152\% & 1888\% & \textbf{2256\%}\\
Double Dunk & -350\% & 669\% & 981\% & \textbf{2000\%} & 1169\%\\
Enduro & 68\% & 164\% & 158\% & 233\% & \textbf{239\%}\\
Fishing Derby & 91\% & 98\% & 98\% & \textbf{106\%} & 105\%\\
Freeway & 101\% & 111\% & 113\% & \textbf{113\%} & 109\%\\
Frostbite & 2\% & 5\% & 33\% & \textbf{83\%} & 69\%\\
Gopher & 120\% & 836\% & 728\% & 1679\% & \textbf{2792\%}\\
Gravitar & -1\% & \textbf{4\%} & -2\% & 1\% & -1\%\\
H.E.R.O. & 47\% & 56\% & 55\% & \textbf{80\%} & 78\%\\
Ice Hockey & 58\% & 58\% & 71\% & \textbf{93\%} & 85\%\\
James Bond 007 & 94\% & 311\% & 161\% & \textbf{1172\%} & 1038\%\\
Kangaroo & 98\% & 339\% & 421\% & \textbf{458\%} & 384\%\\
Krull & 283\% & \textbf{1051\%} & 590\% & 598\% & 653\%\\
Kung-Fu Master & 56\% & 97\% & 146\% & \textbf{153\%} & 151\%\\
Montezuma's Revenge & 1\% & 0\% & 0\% & \textbf{1\%} & -0\%\\
Ms.\ Pac-Man & 4\% & 5\% & 7\% & \textbf{11\%} & 11\%\\
Name This Game & 73\% & 138\% & 143\% & 173\% & \textbf{200\%}\\
Phoenix &  & 270\% & 202\% & 284\% & \textbf{474\%}\\
Pitfall! &  & 2\% & 3\% & -1\% & \textbf{5\%}\\
Pong & 102\% & 110\% & \textbf{111\%} & 110\% & 110\%\\
Private Eye & -1\% & \textbf{2\%} & -2\% & 0\% & -1\%\\
Q*bert & 37\% & \textbf{106\%} & 91\% & 82\% & 93\%\\
River Raid & 25\% & 70\% & 74\% & 81\% & \textbf{128\%}\\
Road Runner & 136\% & \textbf{854\%} & 643\% & 780\% & 850\%\\
Robotank & 863\% & 752\% & \textbf{872\%} & 828\% & 815\%\\
Seaquest & 6\% & 29\% & 36\% & 63\% & \textbf{97\%}\\
Skiing &  & -122\% & 33\% & \textbf{44\%} & 38\%\\
Solaris &  & -21\% & -14\% & \textbf{3\%} & 2\%\\
Space Invaders & 99\% & 118\% & 191\% & 291\% & \textbf{693\%}\\
Stargunner & 378\% & 660\% & 653\% & \textbf{689\%} & 580\%\\
Surround &  & 29\% & 77\% & \textbf{103\%} & 58\%\\
Tennis & 130\% & 130\% & 93\% & 110\% & \textbf{132\%}\\
Time Pilot & 100\% & 89\% & 140\% & 113\% & \textbf{176\%}\\
Tutankham & 16\% & \textbf{67\%} & 63\% & 35\% & 17\%\\
Up'n Down & 28\% & 173\% & 200\% & 125\% & \textbf{313\%}\\
Venture & 4\% & 9\% & 0\% & 7\% & \textbf{22\%}\\
Video Pinball & -5\% & 4042\% & 7221\% & 5727\% & \textbf{7367\%}\\
Wizard of Wor & -15\% & 52\% & 144\% & 131\% & \textbf{177\%}\\
Yars' Revenge &  & \textbf{11\%} & 10\% & 7\% & 10\%\\
Zaxxon & 4\% & 68\% & 102\% & 113\% & \textbf{113\%}\\
\hline
\end{tabular}


} 
} 

\caption{
\label{tab-atari-normalized-30human}
Normalized scores on 57 Atari games (random is $0\%$, human is $100\%$), from a single training run each, using human starts evaluation (see Section~\ref{sec-evaluation}).
Baselines are from~\citet{double-dqn}, see Equation~\ref{eq-norm-score} for how normalized scores are calculated.
}
\end{table*}

\begin{table*}[p]
\centering
\makebox[\textwidth][c]{
\setlength{\tabcolsep}{1.25pt}
{\renewcommand{\arraystretch}{0.8}


\begin{tabular}{|l|rr|rrr|rrr|}
\hline
& \multicolumn{2}{c|}{} & \multicolumn{3}{c|}{DQN} & \multicolumn{3}{c|}{Double DQN (tuned)}\\
\hline
Game & random & human & baseline & Gorila & rank-b. & baseline & rank-b. & prop.\\
\hline
Alien  & 128.3 & 6371.3 & 570.2 & 813.5 & 1191.0 & 1033.4 & \textbf{1334.7} & 900.5\\
Amidar  & 11.8 & 1540.4 & 133.4 & 189.2 & 98.9 & 169.1 & 129.1 & \textbf{218.4}\\
Assault  & 166.9 & 628.9 & 3332.3 & 1195.8 & 3081.3 & 6060.8 & 6548.9 & \textbf{7748.5}\\
Asterix  & 164.5 & 7536.0 & 124.5 & 3324.7 & 9199.5 & 16837.0 & 22484.5 & \textbf{31907.5}\\
Asteroids  & 871.3 & 36517.3 & 697.1 & 933.6 & 1677.2 & 1193.2 & \textbf{1745.1} & 1654.0\\
Atlantis  & 13463.0 & 26575.0 & 76108.0 & \textbf{629166.5} & 207526.0 & 319688.0 & 330647.0 & 593642.0\\
Bank Heist  & 21.7 & 644.5 & 176.3 & 399.4 & 823.7 & \textbf{886.0} & 876.6 & 816.8\\
Battlezone  & 3560.0 & 33030.0 & 17560.0 & 19938.0 & 22250.0 & 24740.0 & 25520.0 & \textbf{29100.0}\\
Beam Rider  & 254.6 & 14961.0 & 8672.4 & 3822.1 & 12041.9 & 17417.2 & \textbf{31181.3} & 26172.7\\
Berzerk  & 196.1 & 2237.5 &  &  & 644.0 & 1011.1 & 865.9 & \textbf{1165.6}\\
Bowling  & 35.2 & 146.5 & 41.2 & 54.0 & 58.0 & \textbf{69.6} & 52.0 & 65.8\\
Boxing  & -1.5 & 9.6 & 25.8 & \textbf{74.2} & 69.6 & 73.5 & 72.3 & 68.6\\
Breakout  & 1.6 & 27.9 & 303.9 & 313.0 & \textbf{481.1} & 368.9 & 343.0 & 371.6\\
Centipede  & 1925.5 & 10321.9 & 3773.1 & \textbf{6296.9} & 2959.4 & 3853.5 & 3489.1 & 3421.9\\
Chopper Command  & 644.0 & 8930.0 & 3046.0 & 3191.8 & \textbf{6685.0} & 3495.0 & 4635.0 & 6604.0\\
Crazy Climber  & 9337.0 & 32667.0 & 50992.0 & 65451.0 & 109337.0 & 113782.0 & 127512.0 & \textbf{131086.0}\\
Defender  & 1965.5 & 14296.0 &  &  & 20634.0 & \textbf{27510.0} & 23666.5 & 21093.5\\
Demon Attack  & 208.3 & 3442.8 & 12835.2 & 14880.1 & 19478.8 & 69803.4 & 61277.5 & \textbf{73185.8}\\
Double Dunk  & -16.0 & -14.4 & -21.6 & -11.3 & -5.3 & -0.3 & \textbf{16.0} & 2.7\\
Enduro  & -81.8 & 740.2 & 475.6 & 71.0 & 1265.6 & 1216.6 & 1831.0 & \textbf{1884.4}\\
Fishing Derby  & -77.1 & 5.1 & -2.3 & 4.6 & 3.5 & 3.2 & \textbf{9.8} & 9.2\\
Freeway  & 0.1 & 25.6 & 25.8 & 10.2 & 28.4 & 28.8 & \textbf{28.9} & 27.9\\
Frostbite  & 66.4 & 4202.8 & 157.4 & 426.6 & 288.7 & 1448.1 & \textbf{3510.0} & 2930.2\\
Gopher  & 250.0 & 2311.0 & 2731.8 & 4373.0 & 17478.2 & 15253.0 & 34858.8 & \textbf{57783.8}\\
Gravitar  & 245.5 & 3116.0 & 216.5 & \textbf{538.4} & 351.0 & 200.5 & 269.5 & 218.0\\
H.E.R.O.  & 1580.3 & 25839.4 & 12952.5 & 8963.4 & 15150.9 & 14892.5 & \textbf{20889.9} & 20506.4\\
Ice Hockey  & -9.7 & 0.5 & -3.8 & -1.7 & -3.8 & -2.5 & \textbf{-0.2} & -1.0\\
James Bond 007  & 33.5 & 368.5 & 348.5 & 444.0 & 1074.5 & 573.0 & \textbf{3961.0} & 3511.5\\
Kangaroo  & 100.0 & 2739.0 & 2696.0 & 1431.0 & 9053.0 & 11204.0 & \textbf{12185.0} & 10241.0\\
Krull  & 1151.9 & 2109.1 & 3864.0 & 6363.1 & \textbf{11209.5} & 6796.1 & 6872.8 & 7406.5\\
Kung-Fu Master  & 304.0 & 20786.8 & 11875.0 & 20620.0 & 20181.0 & 30207.0 & \textbf{31676.0} & 31244.0\\
Montezuma's Revenge  & 25.0 & 4182.0 & 50.0 & \textbf{84.0} & 44.0 & 42.0 & 51.0 & 13.0\\
Ms.\ Pac-Man  & 197.8 & 15375.0 & 763.5 & 1263.0 & 964.7 & 1241.3 & \textbf{1865.9} & 1824.6\\
Name This Game  & 1747.8 & 6796.0 & 5439.9 & 9238.5 & 8738.5 & 8960.3 & 10497.6 & \textbf{11836.1}\\
Phoenix  & 1134.4 & 6686.2 &  &  & 16107.8 & 12366.5 & 16903.6 & \textbf{27430.1}\\
Pitfall!  & -348.8 & 5998.9 &  &  & -193.7 & -186.7 & -427.0 & \textbf{-14.8}\\
Pong  & -18.0 & 15.5 & 16.2 & 16.7 & 18.7 & \textbf{19.1} & 18.9 & 18.9\\
Private Eye  & 662.8 & 64169.1 & 298.2 & \textbf{2598.6} & 2202.3 & -575.5 & 670.7 & 179.0\\
Q*bert  & 183.0 & 12085.0 & 4589.8 & 7089.8 & \textbf{12740.5} & 11020.8 & 9944.0 & 11277.0\\
River Raid  & 588.3 & 14382.2 & 4065.3 & 5310.3 & 10205.5 & 10838.4 & 11807.2 & \textbf{18184.4}\\
Road Runner  & 200.0 & 6878.0 & 9264.0 & 43079.8 & \textbf{57207.0} & 43156.0 & 52264.0 & 56990.0\\
Robotank  & 2.4 & 8.9 & 58.5 & \textbf{61.8} & 51.3 & 59.1 & 56.2 & 55.4\\
Seaquest  & 215.5 & 40425.8 & 2793.9 & 10145.9 & 11848.8 & 14498.0 & 25463.7 & \textbf{39096.7}\\
Skiing  & -15287.4 & -3686.6 &  &  & -29404.3 & -11490.4 & \textbf{-10169.1} & -10852.8\\
Solaris  & 2047.2 & 11032.6 &  &  & 134.6 & 810.0 & \textbf{2272.8} & 2238.2\\
Space Invaders  & 182.6 & 1464.9 & 1449.7 & 1183.3 & 1696.9 & 2628.7 & 3912.1 & \textbf{9063.0}\\
Stargunner  & 697.0 & 9528.0 & 34081.0 & 14919.2 & 58946.0 & 58365.0 & \textbf{61582.0} & 51959.0\\
Surround  & -9.7 & 5.4 &  &  & -5.3 & 1.9 & \textbf{5.9} & -0.9\\
Tennis  & -21.4 & -6.7 & -2.3 & \textbf{-0.7} & -2.3 & -7.8 & -5.3 & -2.0\\
Time Pilot  & 3273.0 & 5650.0 & 5640.0 & \textbf{8267.8} & 5391.0 & 6608.0 & 5963.0 & 7448.0\\
Tutankham  & 12.7 & 138.3 & 32.4 & \textbf{118.5} & 96.5 & 92.2 & 56.9 & 33.6\\
Up'n Down  & 707.2 & 9896.1 & 3311.3 & 8747.7 & 16626.5 & 19086.9 & 12157.4 & \textbf{29443.7}\\
Venture  & 18.0 & 1039.0 & 54.0 & \textbf{523.4} & 110.0 & 21.0 & 94.0 & 244.0\\
Video Pinball  & 20452.0 & 15641.1 & 20228.1 & 112093.4 & 214925.3 & 367823.7 & 295972.8 & \textbf{374886.9}\\
Wizard of Wor  & 804.0 & 4556.0 & 246.0 & \textbf{10431.0} & 2755.0 & 6201.0 & 5727.0 & 7451.0\\
Yars' Revenge  & 1476.9 & 47135.2 &  &  & \textbf{6626.7} & 6270.6 & 4687.4 & 5965.1\\
Zaxxon  & 475.0 & 8443.0 & 831.0 & 6159.4 & 5901.0 & 8593.0 & 9474.0 & \textbf{9501.0}\\
\hline
\end{tabular}


} 
} 

\caption{
\label{tab-atari-raw-30human}
Raw scores obtained on the original 49 Atari games plus 8 additional games where available, evaluated on human starts.  Human, random, DQN and tuned Double DQN scores are from~\citet{double-dqn}.  Note that the Gorila results from~\citet{gorila} used much more data and computation, but the other methods are more directly comparable to each other in this respect.
}

\end{table*}


\end{document}